\documentclass[10pt,twocolumn,letterpaper]{article}
\usepackage[pagenumbers]{cvpr}
\usepackage{times}
\usepackage{graphicx}
\usepackage{amsmath,amssymb}
\usepackage{booktabs}
\usepackage{xspace}

\AtBeginDocument{%
  \captionsetup{aboveskip=3pt,belowskip=1pt}%
  \apptocmd{\normalsize}{%
    \setlength{\abovedisplayskip}{4pt plus 1pt minus 2pt}%
    \setlength{\belowdisplayskip}{4pt plus 1pt minus 2pt}%
    \setlength{\abovedisplayshortskip}{2pt plus 1pt}%
    \setlength{\belowdisplayshortskip}{3pt plus 1pt minus 1pt}}{}{}%
  \normalsize}

\newcommand{\dejaview}{D\'{e}j\`{a} View\xspace}
\newcommand{\ours}{D\'{e}j\`{a} Vite\xspace}

\newcommand{\NA}{\textemdash}

\definecolor{cvprblue}{rgb}{0.21,0.49,0.74}
\usepackage[breaklinks,colorlinks,allcolors=cvprblue]{hyperref}
\usepackage[capitalize]{cleveref}
\def\confName{CVPR}
\def\confYear{2027}
\title{Where to Approximate: Understanding and Accelerating Looped Multi-View Reconstruction}
\hypersetup{pdfauthor={Siyuan Zhang, Yuhan Chen, Yangyang Ning, Nan Wang, Jun Yan}}
\author{%
  Siyuan Zhang\textsuperscript{2}\quad Yuhan Chen\textsuperscript{3}\quad
  Yangyang Ning\textsuperscript{4}\quad Nan Wang\textsuperscript{2,*}\quad
  Jun Yan\textsuperscript{1,*}\\[3pt]
  {\normalsize \textsuperscript{1}Shanghai Ocean University\qquad
   \textsuperscript{2}Independent Researcher}\\
  {\normalsize \textsuperscript{3}Virginia Tech\qquad
   \textsuperscript{4}Tongji University}\\[2pt]
  {\small \href{mailto:homura12138@gmail.com}{\texttt{homura12138@gmail.com}}\quad
   \href{mailto:yuhan@vt.edu}{\texttt{yuhan@vt.edu}}\quad
   \href{mailto:neoyning@tongji.edu.cn}{\texttt{neoyning@tongji.edu.cn}}}\\
  {\small \textsuperscript{*}Corresponding authors:\quad
   \href{mailto:wangnan411570@gmail.com}{\texttt{wangnan411570@gmail.com}}\quad
   \href{mailto:yanjun@ieee.org}{\texttt{yanjun@ieee.org}}}%
}

\begin{document}
\maketitle

\begin{abstract}
Looped multi-view reconstruction refines scene features by repeatedly
applying a shared transformer block. Making these updates cheaper
requires knowing how an approximation at one step affects the rest of
the refinement. We study this with paired interventions in \dejaview
and stacked reconstruction transformers, separating the error an
approximation injects, how that error propagates, and how it affects
the output. In \dejaview, approximation errors are attenuated more
strongly than matched random perturbations under a scale-adjusted
state-error measure, and a local Jacobian decomposition separates the
contribution of channel gating from the response to spatial error
structure. Spatial approximations made early benefit from subsequent
refinement, whereas cross-step reuse depends on which quantity is
cached and how reuse steps are spaced. At equal final norm,
approximation errors change the predicted rotations 1.55 times as much
as random errors. These findings guide \ours, which
reduces spatial resolution and pools keys and values early in
refinement, then interleaves attention-output reuse with full
attention, and is adapted through lightweight self-distillation. Across five
reconstruction benchmarks, the balanced configuration runs
$1.63\times$ faster than the official 16-step \dejaview model with a
1.86-point reduction in average pose AUC@$3^\circ$. At matched latency,
it exceeds the interpolated curve of a trained baseline using full
refinement blocks by 4.73 points in AUC@$3^\circ$ and improves the
corresponding geometry metrics.
\end{abstract}

\section{Introduction}
\label{sec:introduction}
\begin{figure*}[t]
  \centering
  \includegraphics[width=0.74\textwidth]{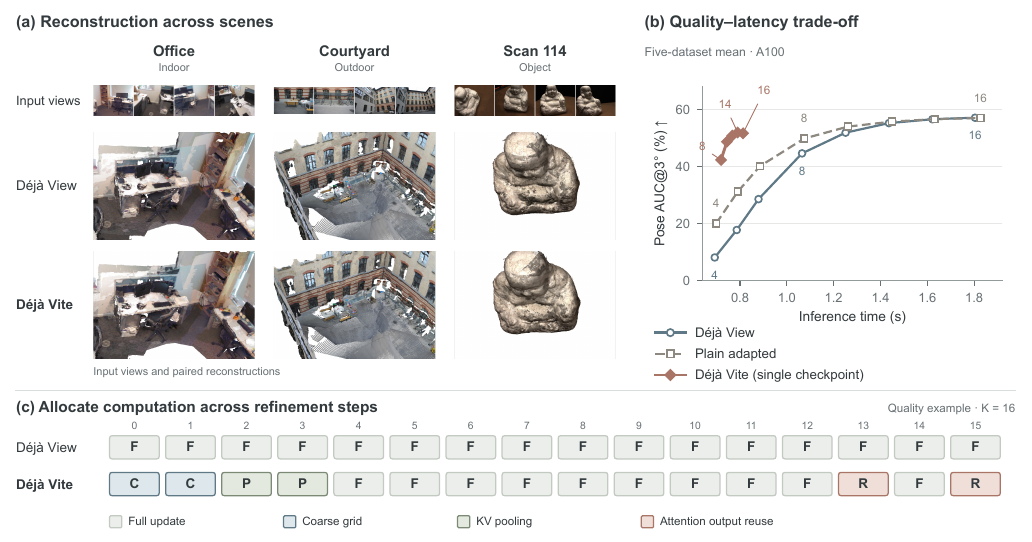}
  \caption{Overview of looped reconstruction and computation allocation.
  (a) Input views and reconstructions from \dejaview and \ours for
  Office, Courtyard, and Scan 114. (b) Pose accuracy versus inference
  time, averaged equally over five datasets on an A100. All measured
  budget settings are plotted; selected markers are labelled with $K$.
  The \ours curve uses a single adapted checkpoint. (c) An example 16-step
  quality schedule, compared with full computation at every step.
  C: coarse-grid processing; P: KV pooling; F: full update;
  R: attention-output reuse.}
  \label{fig:teaser}
\end{figure*}

Feed-forward reconstruction models such as VGGT~\cite{wang2025vggt},
$\pi^3$~\cite{wang2025pi3}, and Depth Anything 3~\cite{lin2025da3}
recover camera poses and dense geometry from image collections with a
deep stack of transformer layers. \dejaview~\cite{burzio2026dejaview}
replaces the stack with a loop: a single shared block refines the scene
features repeatedly, and the number of iterations sets the inference
budget. Spending less then means running fewer iterations, each at
full cost. The loop permits a finer choice (\cref{fig:teaser} (c)), because
an individual iteration can itself be made cheaper, so computation can
be distributed unevenly along the refinement.

The tools for cheaper iterations already exist. Token
merging~\cite{bolya2023tome,shen2025fastvggt} runs a transformer on a
coarser grid with fewer tokens. AVGGT~\cite{sun2026avggt} keeps every
query but lets it attend to a subsampled set of keys and values.
Diffusion models reuse attention outputs from an earlier denoising
step instead of recomputing
them~\cite{ma2024deepcache,liu2025teacache,selvaraju2024fora}. What
these works do not provide is a rule for a loop. Stacked models place
approximations by the role of each layer, and every step of a loop has
the same weights. Diffusion caching judges a reuse by how much the
features changed since the last step, which does not say what the
following steps do with the error.

The cost of an approximate step is not settled at the step itself.
Its error enters a state that the remaining iterations keep refining,
so the cost depends on how those iterations transform the error and on
how much of it reaches the predicted cameras and geometry. We measure
this directly. We run the model twice on the same input, once exactly
and once with a single step approximated, and track the difference
between the two runs through the remaining iterations and the
prediction heads.

Our empirical diagnosis reveals why layer-wise acceleration does not transfer directly to looped models. In a stacked model, each layer has its own weights and its own job,
and accelerations choose what to approximate from that job. A loop
gives no such guide: every step runs the same block, and removing any
single step barely changes the result. What differs between steps is
how much refinement follows them. In a 16-step loop, an error
introduced at step 2 passes through fourteen more updates, and one
introduced at step 14 through two. Two findings show how this decides
where each approximation goes. First, later iterations absorb approximation
errors. A coarse-grid step at iteration 7 costs 21.9 points of
AUC@$3^\circ$ one step later but 2.5 points at the output, whereas the
same step at iteration 12 still costs 10.3 points at the output
(\cref{fig:task_recovery} in the appendix). The loop shrinks these
errors more than random perturbations of equal magnitude, which we
trace to its channel gates: they scale down the channels that carry
the error (\cref{app:gates}). \emph{Spatial approximations therefore
belong early.} Second, reuse errors compound. A reuse step reads an
attention output computed from an earlier state, so each reuse
inherits the drift of those before it: with five reuse steps in a
schedule, the error that the last one makes at its own step is 11.1
times that of the same reuse applied alone. \emph{Reuse is therefore
kept sparse and separated by exact steps.}

Building on these empirical findings, we propose \ours, which runs \dejaview with the following schedule: coarse-grid steps first, KV-pooled steps next, and exact steps afterwards, the last of which alternates with reuse of the global attention output. Low-rank
self-distillation adapts the model to this schedule, and one set of
adapted weights serves several operating points that trade accuracy
for speed. On five benchmarks, our balanced
configuration runs $1.63\times$ faster than the official 16-step
\dejaview model at a cost of 1.86 points in average AUC@$3^\circ$. At
equal latency, it exceeds a trained baseline that keeps full blocks and
lowers the iteration count by 4.73 points and improves the geometry
metrics. With operators and training held fixed, our placement
outperforms uniform placement by 2.62 points.

Our contributions are as follows:
\begin{itemize}
  \item We identify the steps of \dejaview  as interchangeable refinements with a stationary attention pattern, a property enabled by weight sharing and absent from stacked models. Later steps also absorb errors from approximate operations more effectively than random perturbations of equal magnitude. 
  \item We characterize when approximation and reuse are worthwhile: spatial approximations are cheapest early, an error’s output cost depends on its direction, and reuse cost is governed by error compounding, which in turn depends on the cached quantity.
  \item \ours, our acceleration method, runs $1.63\times$ faster than the 16-step model and exceeds a
  trained full-block baseline by 4.73 points in AUC@$3^\circ$ at equal
  latency.
\end{itemize}

\section{Related Work}
\label{sec:related}

\noindent\textbf{Feed-forward reconstruction and its acceleration.}
DUSt3R~\cite{wang2024dust3r} formulates pairwise reconstruction as
pointmap regression, MASt3R~\cite{leroy2024mast3r} adds dense
matching, and MV-DUSt3R+~\cite{tang2025mvdust3r} extends it to
multiple views in a single pass. VGGT~\cite{wang2025vggt},
$\pi^3$~\cite{wang2025pi3}, and Depth Anything 3~\cite{lin2025da3}
process all views jointly with stacked transformers, and their global
attention dominates the cost for large image sets. Training-free
accelerations reduce this cost. FastVGGT~\cite{shen2025fastvggt}
merges tokens following ToMe~\cite{bolya2023tome}, block-sparse global
attention~\cite{wang2026blocksparse} keeps the concentrated cross-view
interactions, and AVGGT~\cite{sun2026avggt} converts early global
layers to frame attention and subsamples keys and values in the rest.
These methods decide where to approximate from the roles of distinct
layers. A looped model has a single block, so the decision becomes
when to approximate, and we derive it from how errors propagate.

\noindent\textbf{Looped computation and its dynamics.}
Universal Transformers~\cite{dehghani2019universal} and
ALBERT~\cite{lan2020albert} share weights across depth, deep
equilibrium models~\cite{bai2019deq} take this to a fixed-point limit,
and looped transformers learn iterative
algorithms~\cite{yang2024looped} whose stability has been studied
through their fixed points~\cite{labovich2026stability}. In geometric
vision, RAFT~\cite{teed2020raft} and
DROID-SLAM~\cite{teed2021droid} refine estimates with a shared
recurrent update. \dejaview~\cite{burzio2026dejaview} brings looping
to multi-view reconstruction and shows that its state reaches no fixed
point: the feature norm grows while the direction stabilizes.
Raptor~\cite{jacobs2026raptor} rewrites vision transformers with a few
recurrent blocks and observes trajectories that self-correct under
small random perturbations. Concurrently,
Solv3R~\cite{kwak2026solv3r} makes the \dejaview update independent of
the iteration index and tests its stability under large random noise
and state resets. We study the errors produced by
approximate operations and show that they propagate differently from
random perturbations of equal magnitude and affect the decoded output
differently.

\noindent\textbf{Cross-step reuse.}
Diffusion models reuse computation across denoising steps.
DeepCache~\cite{ma2024deepcache} reuses slowly varying high-level
features, TeaCache~\cite{liu2025teacache} selects refresh steps from
timestep-modulated inputs, and FORA~\cite{selvaraju2024fora},
PAB~\cite{zhao2025pab}, and T-GATE~\cite{liu2025tgate} reuse attention
outputs. Later methods target the error that caching accumulates by
correcting it~\cite{qiu2025eoc,qiu2025goc}, searching non-uniform
caching patterns~\cite{cao2026procache}, or forecasting
features~\cite{liu2025taylorseer}. We study reuse inside a
deterministic refinement loop and show that accumulation depends on
the cached quantity, which sets the spacing of reuse steps.

\noindent\textbf{Our position.}
\dejaview can save computation only by cutting loop iterations. Guided
by our findings, \ours instead makes individual iterations cheaper
and places each approximation by how its error propagates.

\section{Understanding Errors in the Loop}
\label{sec:analysis}
\label{sec:prelim}
\label{sec:mechanism_experiments}

\subsection{Setup}
\label{sec:looped_model}
\label{sec:protocol}
\label{sec:local_model}
Given images $\mathcal I$, \dejaview initializes $z_0=E(\mathcal I)$
and applies a shared, time-conditioned block $K$ times
\cite{burzio2026dejaview}:
\begin{equation}
  z_{t+1}=f_\theta(z_t;\tau_t,\tau_{t+1}),\qquad
  \tau_t=\min\!\left(\frac{t}{K-1},1\right).
  \label{eq:recurrence}
\end{equation}
The block contains a frame-attention and a global-attention subblock,
each with attention and a feed-forward layer. Each subblock ends with a
channel-wise scaling of its output, which we call a channel gate.
Prediction heads decode $z_K$ into camera poses and geometry.

We study three approximate operations, defined in \cref{sec:operators}.
A \emph{coarse-grid} step runs the whole block on $2\times2$
average-pooled tokens. A \emph{KV-pooling} step keeps all queries and
pools the keys and values of global attention over $2\times2$ windows.
A \emph{reuse} step replaces the global-attention output with the one
computed at the preceding step.

\noindent\textbf{Paired interventions.}
For each scene we run a clean trajectory and a perturbed trajectory
with identical inputs. At step $u$ we replace one operation and then
resume exact computation. Let $\delta_t=\widetilde z_t-z_t$ be the
state deviation. Because the state norm grows along the trajectory, we
measure the scale-adjusted deviation $\widehat A_t=R_tm_t$, the product
of the mean relative token deviation $R_t$ and the mean clean token
norm $m_t$, and report the retained fraction
$r_u=\widehat A_K/\widehat A_{\mathrm{inj}}$ from injection to output.
At the output we measure the rotation
deviation from the clean prediction and the change in pose AUC and
AbsRel against ground truth.

We intervene at steps 2, 7, and 12 of the 16-step \dejaview model and
at 15\%, 50\%, and 85\% of depth in the released
VGGT~\cite{wang2025vggt}, $\pi^3$~\cite{wang2025pi3}, and
DA3~\cite{lin2025da3}. The analysis uses 65 scenes, 13 from each of
five datasets, with at most 32 views. Propagation measurements use FP32.
Placement rules are derived on the 35 even-indexed scenes and tested on
the 30 odd-indexed scenes. \Cref{app:mechanism} gives the full protocol
and additional analyses. Figures and tables numbered with the prefix S are
in the appendix.

\subsection{Loop Steps Are Interchangeable Refinements}
\label{sec:loop_properties}
Before approximating steps, we ask what a step does. Three properties
separate the loop from stacked models (\cref{tab:loop_properties}).

\begin{table}[t]
  \centering
  \small
  \setlength{\tabcolsep}{3pt}
  \caption{Properties of the loop on 65 scenes. Skip and cross-view
  ablation report the scene median of the AUC@$30^\circ$ drop, in
  points, at the worst step or layer of each scene. The last column
  counts scenes whose global attention pattern is stationary over the
  final units.}
  \label{tab:loop_properties}
  \begin{tabular}{lrrr}
    \toprule
    Model & Worst skip & Worst cross-view & Stationary \\
          &            & ablation         & attention \\
    \midrule
    \dejaview & 0.18 & 1.9 & 65/65 \\
    VGGT & 48.8 & 15.9 & 0/65 \\
    $\pi^3$ & 54.8 & 11.9 & 0/65 \\
    DA3 & 82.2 & 7.5 & 0/65 \\
    \bottomrule
  \end{tabular}
\end{table}

\noindent\textbf{No step is critical.}
We skip one update at a time and record the worst step of each scene.
In \dejaview the worst skip costs a median of 0.18 points of
AUC@$30^\circ$ and at most 1.32 points. Every scene of every stacked
model contains a layer whose removal costs more than this maximum,
with medians of 48.8, 54.8, and 82.2 points for VGGT, $\pi^3$, and
DA3. The cross-view read of a single step is just as dispensable.
Replacing the other views' keys and values with those of another scene
costs a median of 1.9 points at the worst step of \dejaview, against
15.9, 11.9, and 7.5 points in the stacked models.

\noindent\textbf{Attention stops changing.}
From 27\% of the loop onward, the global attention patterns of
consecutive steps are nearly identical in all 65 scenes. Over the
trajectory the relative update size falls from 0.76 to 0.07, and
consecutive updates align in direction, with cosine rising from 0.17
to 0.98. No stacked model reaches a stationary pattern in any scene.
Late steps thus repeat one operation with a shrinking step size, which
is what makes a late attention output reusable.

\noindent\textbf{Weight sharing is the cause.}
We fine-tune two eight-step models from the \dejaview block with the
same data, loss, and schedule, one with tied and one with untied
weights. The tied model stays flat: skipping any step costs 3 to 4
points of AUC@$3^\circ$. The untied model specializes its last block,
whose removal costs 18 to 19 points across two seeds, and it is less
accurate, with AUC@$3^\circ$ of 46.2 and 47.7 against 53.2
(\cref{tab:tied_untied}). The
property survives removing the time conditioning and training with a
fixed number of steps.

Since steps are interchangeable refinements, an approximation is
placed by its position along the trajectory, with no layer role to
consult. The remaining subsections study how that position matters.
\Cref{app:loop_properties} gives the protocols and per-step results.

\subsection{Later Iterations Absorb Approximation Errors}
\label{sec:propagation}
\label{sec:gate_mechanism}
\label{sec:jvp_experiments}
\label{sec:output_mechanism}
A single approximation costs less at the output than at the step where
it is applied (\cref{fig:task_recovery}). A coarse-grid step at
iteration 7 costs 21.9 points of AUC@$3^\circ$ one step later and 2.5
points at the output. At iteration 2 all three operations end with no
measurable loss, and at iteration 12 the coarse-grid step still costs
10.3 points.

The state shows the same behavior
(\cref{tab:matched_perturbations}). \dejaview reduces the
scale-adjusted error in all 12 operator--position conditions. It
retains 0.29--0.85 of the error at the output, and each fraction is
below that of a random perturbation of the same magnitude. The
difference comes from the channel gates. Along the path through the
gates alone, a KV-pooling error injected after step 12 has a median
gain of 0.469, against 0.767 for random directions, with the same
ordering in all 65 scenes (\cref{fig:propagation}). Approximation
errors carry their energy in channels that the gates scale down,
whereas random perturbations spread it over all channels. The
attenuation is a contraction by the gates (\cref{app:gates}).

VGGT and DA3 show the opposite preference. VGGT retains more of the
approximation error than of the random perturbation in five of six
conditions, and DA3 in all six. Their absolute retained fractions
exceed one in 17 of 18 conditions (\cref{app:propagation}).

The output error depends on the direction of the remaining state error
as well as its size. When
final residuals are rescaled to equal norm and decoded, the residual
of an approximation changes the predicted rotations 1.55 times as much
as a random residual, and more in 62 of 65 scenes
(\cref{tab:head_direction,app:output}). Decoded pose and geometry are therefore
our criterion for comparing placements and schedules.

\subsection{Spatial Approximations Belong Early}
\label{sec:placement}
\label{sec:composition_model}
Since later iterations absorb errors, spatial approximations are
cheapest early (\cref{fig:placement,app:placement}). With four coarse-grid steps,
the early layout gives a rotation deviation of $0.30^\circ$, against
$1.69^\circ$ for uniform and $1.78^\circ$ for reversed placement, and
it wins on all 30 held-out scenes. With four KV-pooling steps the
deviations are $0.17^\circ$ and $0.32^\circ$, with wins on 29 of 30
scenes. Reuse behaves differently. Moving it from late to early steps
increases the deviation by $0.41^\circ$, and late placement is
preferred in 60 of 65 scenes, because its immediate error is smaller
late in the trajectory.

\Cref{sec:schedule_ablation} confirms this placement after adaptation,
and \cref{app:transfer} reports its transfer across budgets and view
counts. A predictor fitted on these measurements ranks the cost of
held-out operators and budgets (\cref{app:prediction}).

\subsection{Reuse Errors Compound Across Steps}
\label{sec:cache_analysis}
A reuse step depends on an earlier state, so repeated reuse interacts.
With five reuse steps at $\{6,8,10,12,14\}$, the local error of the
last output reuse is $11.1\times$ that of the same reuse applied alone.
The factor is $7.0\times$ for values and $7.5\times$ for key--value
pairs, but $1.6\times$ for keys and $2.0\times$ for attention
probabilities (\cref{fig:cache,app:reuse}). Output reuse should therefore be
spread out. On 30 held-out scenes, spreading five output-reuse steps
lowers the rotation deviation from $0.710^\circ$ to $0.374^\circ$ and
wins on 25 scenes. The stability of the cached quantity predicts the
local error of a single reuse, with Spearman correlation 0.95--1.00.
The final cost is set by how these local errors compound along the
trajectory, so we select reuse layouts by evaluating complete
schedules.

\section{Accelerating Looped Reconstruction}
\label{sec:method}
Our method, \ours, has three parts, each answering one question. The
approximate operations (\cref{sec:operators}) decide \emph{how} a step
is made cheaper. The compute schedule (\cref{sec:schedule}) decides
\emph{when} each operation runs, using the findings of
\cref{sec:analysis}. Adaptation (\cref{sec:adaptation}) recovers the
accuracy that the approximations cost and lets one set of weights
serve several speed settings. We keep the notation of
\cref{eq:recurrence}: $z_t$ is the state before step $t$, $K$ is the
number of steps, and $\tau_t$ is the relative time of step $t$.

\subsection{Approximate Operations}
\label{sec:operators}
\begin{figure*}[t]
  \centering
  \includegraphics[width=0.30\textwidth]{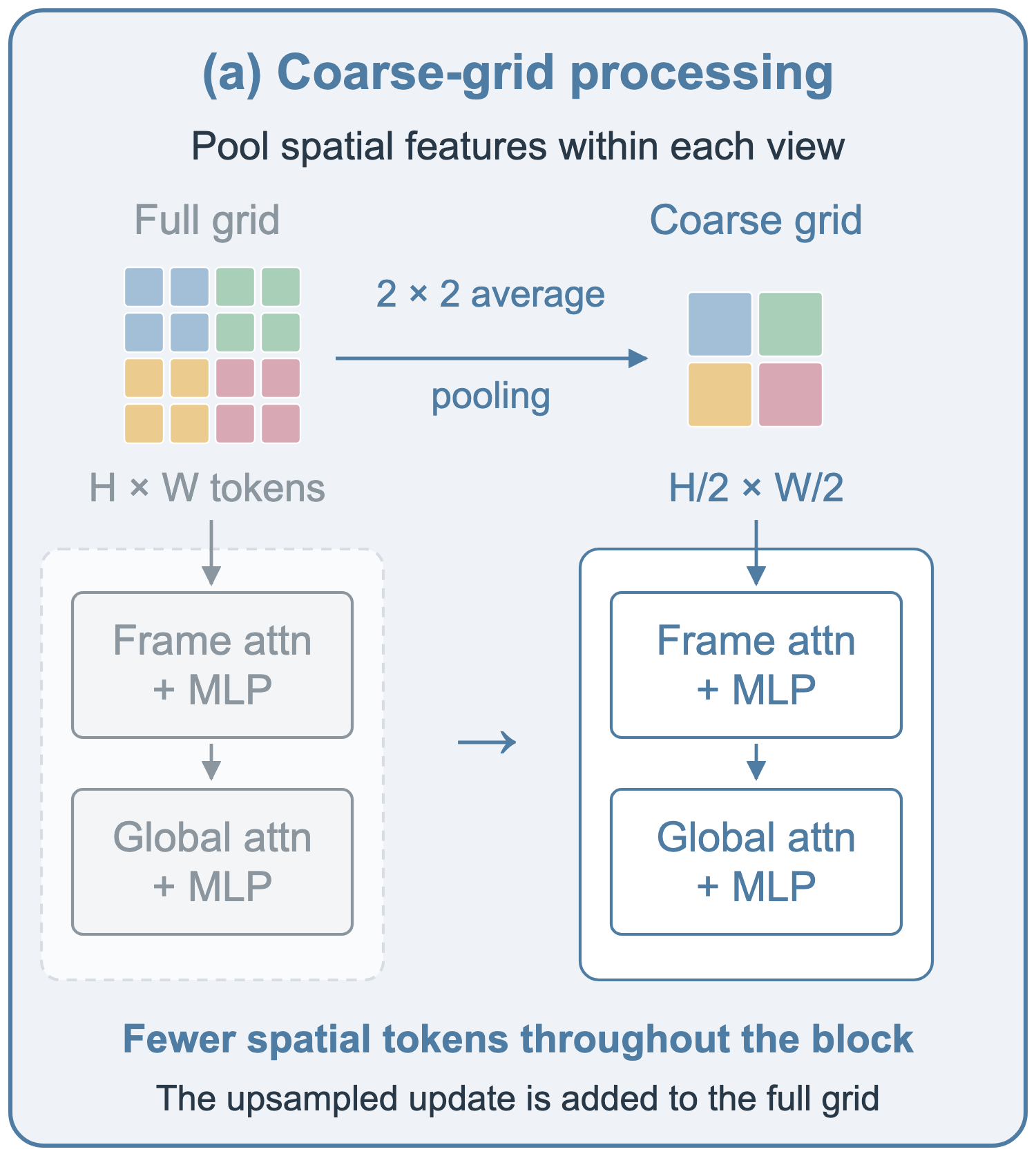}\hfill
  \includegraphics[width=0.30\textwidth]{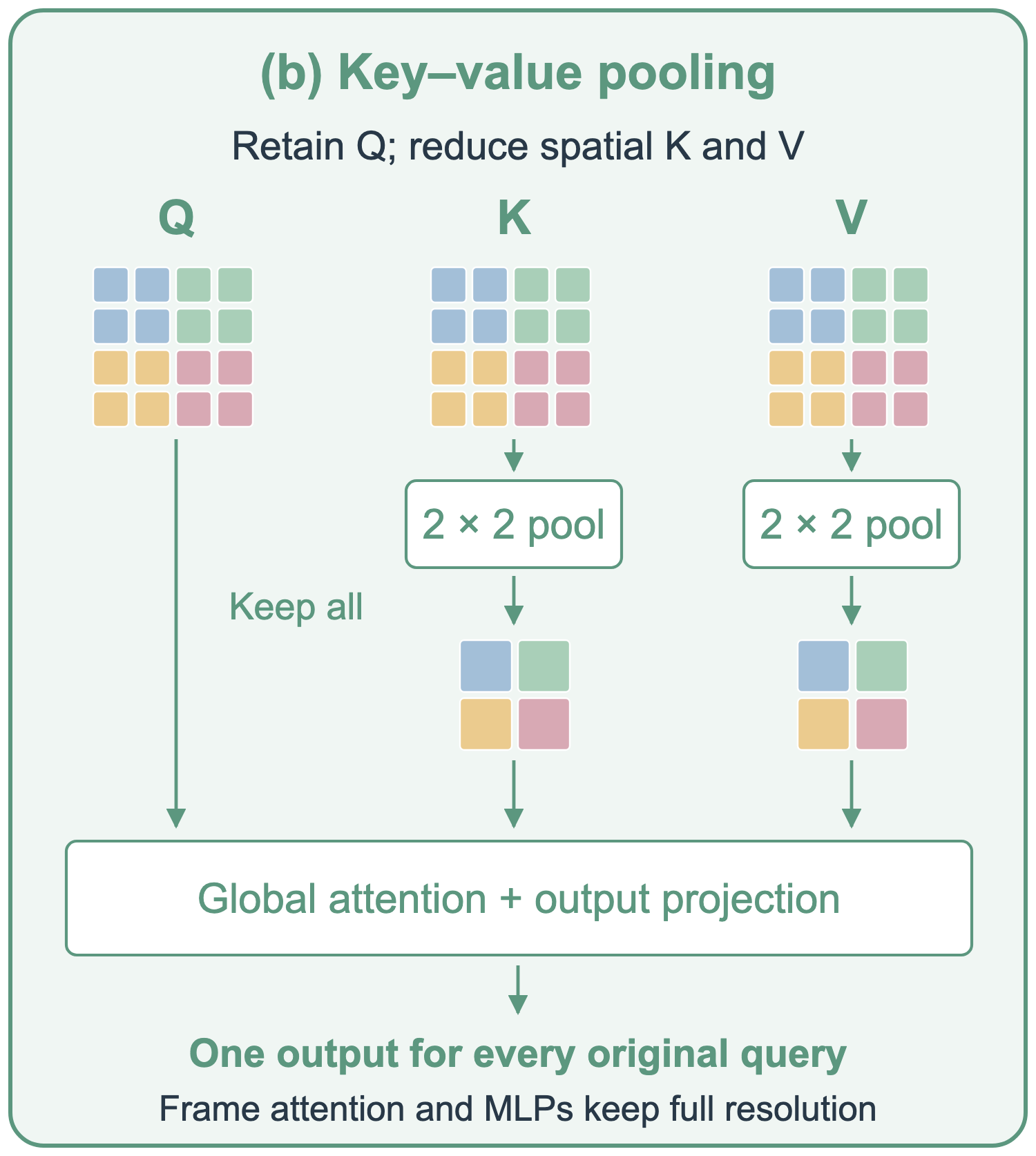}\hfill
  \includegraphics[width=0.30\textwidth]{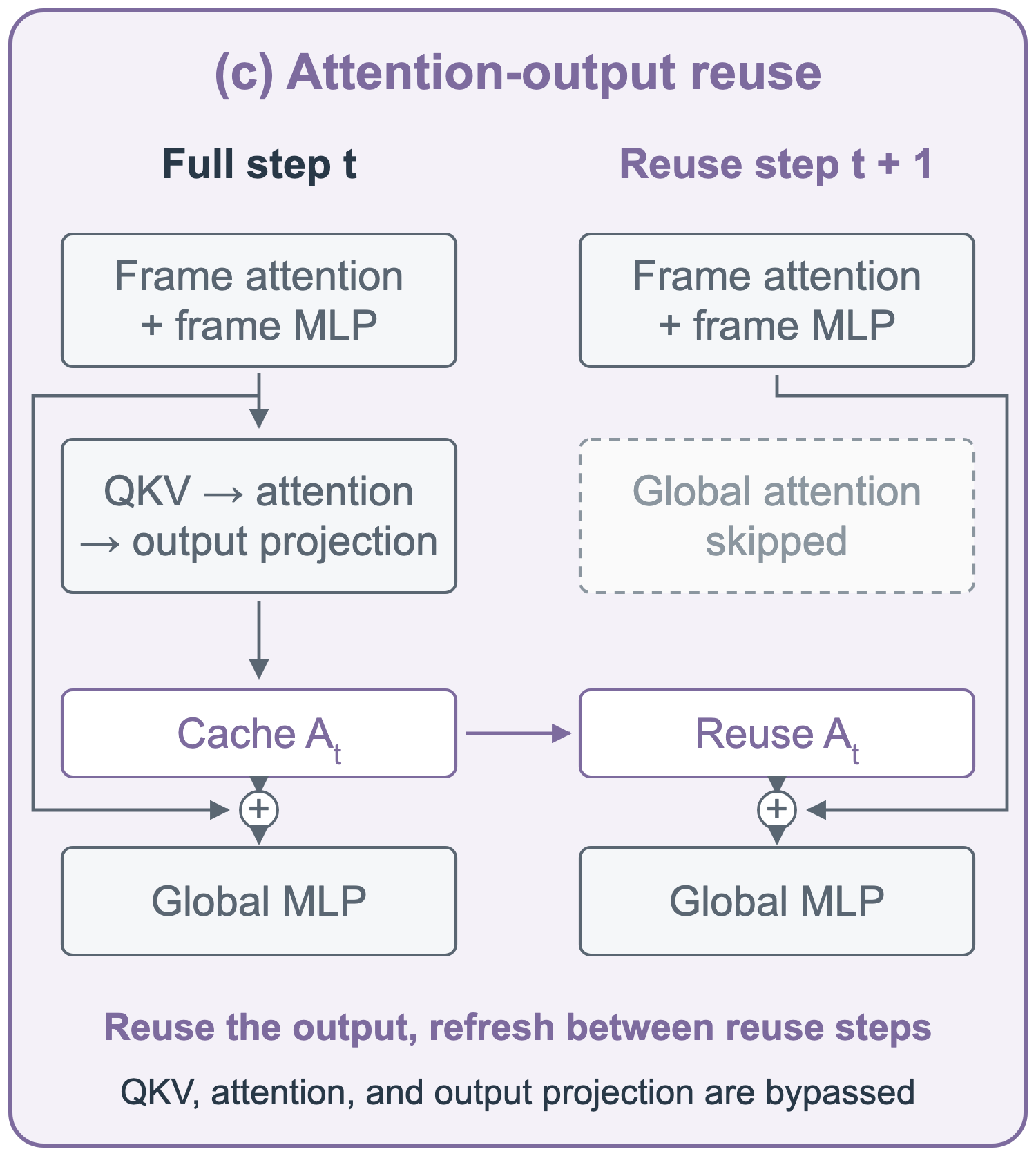}
  \caption{Three approximate operations used by \ours.
  (a) Coarse-grid processing applies $2\times2$ spatial average pooling
  before the entire recurrent block. Both attention subblocks and their
  feed-forward layers process the reduced grid, and the resulting update
  is upsampled and added to the full-resolution state; $H,W$ denote
  patch-grid dimensions here. The gray path shows full-resolution
  computation for comparison. (b) KV pooling retains all queries and
  pools spatial keys and values within each view before global
  attention. The output retains the original query resolution. In (a)
  and (b), colors identify spatial pooling groups and special tokens,
  which are never pooled, are not depicted. (c) A full step caches its projected
  global-attention output $A_t$ for the next reuse step. The dashed block
  is bypassed, while frame attention, both MLPs, and the current-state
  residual path remain active. Normalization, gating, and MLP residual
  connections are omitted from the schematic.}
  \label{fig:method}
\end{figure*}

A step of the loop spends its time in two places: every layer scales
with the number of tokens, and global attention additionally scales
with the number of query--key pairs across all views. The three
operations in \cref{fig:method} remove these costs at different
granularity. Coarse-grid processing shrinks the token set of the whole
block, KV pooling shrinks only the key set of global attention, and
cross-step reuse skips global attention altogether.

\noindent\textbf{Coarse-grid processing.}
This operation is the cheapest of the three and the least precise. It
runs a step at half the spatial resolution, which suits the first
steps of the loop, where the state is still coarse and later steps
absorb the error (\cref{sec:placement}). We construct the coarse token
grid by applying $2\times2$ average pooling to the patch tokens of
each view. Let $X\in\mathbb{R}^{H\times W\times D}$ be the patch
tokens of one view, arranged on the $H\times W$ patch grid with
channel width $D$, and let $X_{i,j}\in\mathbb{R}^{D}$ be the token at
row $i$ and column $j$. The pooling operator $\mathcal{D}_2$ maps $X$
to a grid of size
$\lceil\tfrac{H}{2}\rceil\times\lceil\tfrac{W}{2}\rceil$ whose token
at location $(i,j)$ is
\begin{equation}
  [\mathcal{D}_2(X)]_{i,j}
  =\frac{1}{|\Omega_{i,j}|}\sum_{(r,c)\in\Omega_{i,j}}X_{r,c},
  \label{eq:spatial_pooling}
\end{equation}
where $\Omega_{i,j}$ holds the grid locations $(2i+u,\,2j+v)$,
$u,v\in\{0,1\}$: four tokens, or fewer at the border of an odd-sized
grid. The
entire recurrent block operates on this coarse grid, including
frame attention, global attention, and the feed-forward layers. This
reduces the number of spatial tokens by approximately a factor of four
in a coarse step. The camera and register tokens are not pooled, and
each coarse token takes the center of its window as its position in
the rotary position encoding.

The state itself stays at full resolution. Let $\mathcal{U}$ denote
nearest-neighbor upsampling, which copies each coarse token to the
locations of its window and so restores the $H\times W$ grid. A coarse step updates the patch tokens
as
\begin{equation}
  z_{t+1}=z_t+\mathcal{U}\!\left(
  f_\theta(\mathcal{D}_2 z_t)-\mathcal{D}_2 z_t\right),
  \label{eq:coarse_step}
\end{equation}
where $\mathcal{D}_2$ acts on each view and the time arguments of
$f_\theta$ are omitted. The change that the block makes on the coarse
grid is added to every token of the window, so the detail finer than a
window passes through a coarse step unchanged and the following steps
resume on the full-resolution grid without a separate recovery
operator.

\noindent\textbf{Key--value pooling.}
This operation targets global attention alone, the most expensive
layer for large image sets, and keeps the state at full resolution. It
is the intermediate option between a coarse step and an exact step:
every token is still updated at its own location, but it reads from a
four times smaller set of keys and values. KV pooling retains
full-resolution queries and pools the spatial keys and values over
$2\times2$ neighborhoods within each view, using the operator of
\cref{eq:spatial_pooling}. Let $N$ be the number of tokens over all
views and $d$ the per-head dimension. At step $t$, let
$Q_t\in\mathbb{R}^{N\times d}$ stack the queries of all tokens, and
let $\overline{K}_t$ and $\overline{V}_t$ stack the pooled keys and
values of all views, with about $N/4$ rows each. An overline marks a
pooled quantity. The camera and register tokens are not pooled, and
$K^{\mathrm s}_t$ and $V^{\mathrm s}_t$ stack their keys and values.
For each attention head, the output is
\begin{equation}
  H_t=\operatorname{softmax}\!\left(\left[
  \frac{Q_tK_t^{\mathrm s\mathsf T}}{\sqrt{d}},\;
  \frac{Q_t\overline{K}_t^{\mathsf T}}{\sqrt{d}}+b\right]\right)
  \begin{bmatrix}V^{\mathrm s}_t\\ \overline{V}_t\end{bmatrix},
  \label{eq:pooled_attention}
\end{equation}
where the brackets concatenate the two groups of keys, the softmax
normalizes each row over all keys, and the scalar bias $b$ is added to
the logits of the pooled keys only. Row $n$ of $H_t$ is the attention output for token
$n$. Head outputs are concatenated
and passed through the output projection. Each query therefore receives
an updated feature at its original spatial location, while attending
to a smaller set of keys and values. Frame attention and the
feed-forward layers operate at full resolution in a KV-pooled step.

The pooled keys and values are window means with $b=\log 4$, so that a
pooled key carries the attention mass of the four keys it replaces.
When tokens are constant within each window, the pooled attention
equals full attention. This mean pooling needs no training and is the
form used in the interventions on released models. In the adapted
model, the P steps use \emph{learned KV pooling}: each attention head
has learned window weights that replace the mean and its own learned
bias $b$.
Learned pooling is initialized to mean pooling.

\noindent\textbf{Cross-step reuse.}
This operation removes global attention from a step entirely. It
exploits the stationary attention pattern of the loop
(\cref{sec:loop_properties}): consecutive steps attend to the same
tokens, so the previous output can stand in for the current one. A reuse step
takes the global-attention output from the preceding step. Let $x_t$
be the tokens that enter the global-attention subblock at step $t$,
and let $\mathcal{G}(x_t)$ denote the global-attention computation,
including the query, key, and value projections and the output
projection. Let $\mathcal{S}_{\mathrm{reuse}}\subset\{1,\ldots,K-1\}$
be the set of reuse steps; the first step is never one. The attention output $A_t$ supplied to the
rest of the block is
\begin{equation}
  A_t=
  \begin{cases}
    A_{t-1}, & t\in\mathcal{S}_{\mathrm{reuse}},\\
    \mathcal{G}(x_t), & \text{otherwise}.
  \end{cases}
  \label{eq:attention_reuse}
\end{equation}
A full evaluation computes and caches $A_t$; a reuse step reads the
cache and skips the projections and attention computation. Frame
attention and the feed-forward layers continue to update the current
state at every step. The cache stores one $N\times D$ tensor, one
$D$-dimensional output per token. The compute schedule in \cref{sec:schedule} specifies the
reuse steps.

\subsection{From Findings to a Compute Schedule}
\label{sec:schedule}
The operations above say how to save computation. The schedule says
when, and this is where the analysis enters: the same operation costs
little accuracy at one step and much at another, so the order of the
steps determines the trade-off. We write a schedule as a string with
one letter per step, read from the first step to the last. C, P, F,
and R denote a coarse-grid step, a KV-pooled step, a full step, and a
reuse step. The three operating points use the schedules
$\mathrm{C}^2\mathrm{P}^2\mathrm{F}^9\mathrm{RFR}$ ($K=16$),
$\mathrm{C}^2\mathrm{P}^4\mathrm{F}^5\mathrm{RFR}$ ($K=14$), and
$\mathrm{C}^6\mathrm{P}^4\mathrm{FRFR}$ ($K=14$) for the quality,
balanced, and fast settings. A superscript repeats a step, so
$\mathrm{C}^2\mathrm{P}^4\mathrm{F}^5\mathrm{RFR}$ runs two coarse
steps, four KV-pooled steps, five full steps, and then a reuse step, a
full step, and a reuse step. The string length equals $K$: every step
advances the loop index $t$, including reuse steps.
Spatial approximation occupies the early trajectory, followed by
full-resolution refinement. Interleaving late reuse with full attention
refreshes the cache between reuse steps. This arrangement follows the
position and composition findings in
\cref{sec:placement,sec:cache_analysis};
operator counts set the three operating budgets. Matched-adaptation
schedule comparisons in \cref{sec:schedule_ablation} test the resulting
allocation against uniform, reversed, stability-based, and random
layouts.

\subsection{Adaptation and Budget Control}
\label{sec:adaptation}
The pretrained block was trained to run exact steps, so the scheduled
approximations cost accuracy when applied directly. Adaptation closes
this gap with two requirements. The correction must be able to differ
along the trajectory, because the early steps are coarse and the late
steps reuse attention. It must also be defined for any number of
steps, so that one set of weights serves every schedule. We meet both
by placing a small number of low-rank corrections, which we call
\emph{anchors}, at fixed relative times and interpolating between
them.

We use nine sets of rank-16 LoRA anchors~\cite{hu2021lora}, applied to
the query--key--value and output projections and both feed-forward
projections in the frame and global attention subblocks. Let
$W\in\mathbb{R}^{d_{\mathrm{out}}\times d_{\mathrm{in}}}$ be one of
these pretrained projection matrices, shared by all steps, with input
width $d_{\mathrm{in}}$ and output width $d_{\mathrm{out}}$. Anchor
$j\in\{0,\ldots,8\}$ stores a low-rank update $\Delta W_j$ of $W$ as
two factors:
\begin{equation}
  \begin{aligned}
    \Delta W_j&=U_jV_j,\\
    V_j&\in\mathbb{R}^{16\times d_{\mathrm{in}}},\quad
    U_j\in\mathbb{R}^{d_{\mathrm{out}}\times16}.
  \end{aligned}
  \label{eq:adapter}
\end{equation}
Here $V_j$ projects the input to 16 dimensions and $U_j$ maps it back
to the output width. The adapter scaling is absorbed into the factors.
Anchor $j$ lies at relative time $j/8$, so the nine anchors divide the
trajectory into eight equal intervals. Step $t\in\{0,\ldots,K-1\}$ has
relative time $\tau_t=t/(K-1)$ and falls at position $p_t=8\tau_t$ on
the anchor axis. Let $j_0=\lfloor p_t\rfloor$ and
$j_1=\min(j_0+1,8)$ be the two neighboring anchors and
$w_t=p_t-j_0\in[0,1)$ the fractional distance from the first. We
linearly interpolate the anchor updates:
\begin{equation}
  \Delta W_t^{(K)}=(1-w_t)\,U_{j_0}V_{j_0}
                    +w_t\,U_{j_1}V_{j_1}.
  \label{eq:anchor_interpolation}
\end{equation}
The superscript records that the update depends on the step count $K$
through $\tau_t$. Step $t$ uses the projection $W+\Delta W_t^{(K)}$.
All four step types use this position rule. Because the anchors are
indexed by relative time, the same anchor sets serve every iteration
budget. With feature width 768 and feed-forward width
3072, the nine sets contain 3,538,944 LoRA parameters
(\cref{app:setup}). The block
receives the original time pair $(\tau_t,\tau_{t+1})$ of
\cref{eq:recurrence}, which is $(1,1)$ at the last step.

Adaptation uses label-free self-distillation for 12,000 training steps,
with the official $K=16$ model as the teacher. The student executes
approximate computation during adaptation, including the scheduled
reuse operations. The loss is computed on the outputs decoded from the
final state by the frozen heads: $\ell_1$ distances to the teacher's
depth, point map, ray map, and pose, an angular error on ray
directions, a relative pose error over view pairs, and a term that
fits the pooled attention output to the exact one in KV-pooling steps
(\cref{eq:distill_loss} in \cref{app:setup}). The distillation data follow the training datasets
of \dejaview: we randomly draw about 3\% of the images of their
training splits.
Before
inference, we derive each step's update from the anchors and precompute
its merged projection for execution at the requested budget. The
checkpoint stores the shared base weights and the nine anchor sets.

A single checkpoint can also cover a range of step counts. For the
adapter used in the budget sweep, training samples
$K\in\{8,10,12,14,16\}$. The evaluated budgets are all represented
during training. At inference, a requested budget selects $K$ and
its operator schedule while retaining the original time conditioning.
This checkpoint runs $\mathrm{C}^{K-8}\mathrm{P}^{4}\mathrm{FRFR}$ at
every budget. At $K=14$ this is the schedule of the fast setting; the
two are separately trained checkpoints, one on this schedule family
and one shared by the three operating points
(\cref{app:main_results}).
All budget settings in this sweep use the same
checkpoint containing the base weights and the nine LoRA anchor sets;
changing the budget changes the number and type of
updates applied to the state. The image encoder and prediction heads
retain their original computation.
The tier adapters use the same nine-anchor parameterization with the
operating-point schedules in \cref{sec:schedule}.

Finally, a learned multistep update improves accuracy at low step
counts. Each step applies a per-channel weighted combination of its
own update and the two previous ones, as multistep solvers do for
differential equations (\cref{eq:multistep} in \cref{app:setup}). The
weights follow the same anchor construction as the adapters, with
seven uniformly spaced anchors at relative times $j/6$,
$j\in\{0,\ldots,6\}$, linearly interpolated at $\tau_t$. A run of
coarse steps and its update history stay on the coarse grid, which
equals applying \cref{eq:coarse_step} at every step, so the history
is cleared at each change of resolution.

\section{Experiments}
\label{sec:experiments}
\label{sec:acceleration_experiments}

\subsection{Setup}
\label{sec:setup}
We follow the released \dejaview evaluation
protocol~\cite{burzio2026dejaview} on DTU~\cite{jensen2014dtu},
ETH3D~\cite{schops2017eth3d}, 7-Scenes~\cite{shotton2013scene},
ScanNet++~\cite{yeshwanth2023scannetpp}, and
nuScenes~\cite{caesar2020nuscenes}, 153 scenes in total. We report pose AUC@$3^\circ$ and
AUC@$30^\circ$, and, after aligning the prediction to the ground truth
with a similarity transform (Sim(3)), pointmap AbsRel (the mean
relative $\ell_2$ point error) and inlier ratio. Averages weight the five datasets equally. The reference is
the official 117M-parameter \dejaview checkpoint at $K=16$. We also
evaluate it at lower $K$, and we compare with a trained plain baseline
that keeps full refinement blocks at each $K$
(\cref{app:main_results}). Latency is the median
of five runs after two warmup runs on one A100 with bfloat16
inference, and covers the encoder, the loop, and the prediction heads.
\Cref{app:experiments} gives the full protocol.

\subsection{Main Results}
\label{sec:main_results}
\begin{table*}[t]
  \centering
  \footnotesize
  \setlength{\tabcolsep}{1.4pt}
  \renewcommand{\arraystretch}{0.90}
  \caption{Reconstruction accuracy and inference time on five benchmarks.
  Each dataset reports pose AUC@$3^\circ$ (\%, higher is better),
  AbsRel ($\times 10^3$, lower is better), and time (seconds, lower is better).
  Bold indicates the best value in each column.}
  \label{tab:main}
  \begin{tabular*}{\textwidth}{@{\extracolsep{\fill}}l*{15}{r}@{}}
    \toprule
    & \multicolumn{3}{c}{DTU}
    & \multicolumn{3}{c}{ETH3D}
    & \multicolumn{3}{c}{7-Scenes}
    & \multicolumn{3}{c}{ScanNet++}
    & \multicolumn{3}{c}{nuScenes} \\
    & \multicolumn{3}{c}{49 views}
    & \multicolumn{3}{c}{14--76 views}
    & \multicolumn{3}{c}{46--91 views}
    & \multicolumn{3}{c}{100 views}
    & \multicolumn{3}{c}{39--41 views} \\
    \cmidrule(lr){2-4}\cmidrule(lr){5-7}\cmidrule(lr){8-10}\cmidrule(lr){11-13}\cmidrule(lr){14-16}
    Method & AUC$\uparrow$ & AbsRel$\downarrow$ & Time$\downarrow$
    & AUC$\uparrow$ & AbsRel$\downarrow$ & Time$\downarrow$
    & AUC$\uparrow$ & AbsRel$\downarrow$ & Time$\downarrow$
    & AUC$\uparrow$ & AbsRel$\downarrow$ & Time$\downarrow$
    & AUC$\uparrow$ & AbsRel$\downarrow$ & Time$\downarrow$ \\
    \midrule
    VGGT & \textbf{96.48} & 9.6 & 3.007 & 35.42 & 56.7 & 1.707 & 11.13 & 42.8 & 7.246 & 14.92 & 42.6 & 7.570 & 39.15 & 77.1 & 1.525 \\
    AVGGT & 84.94 & 11.7 & 1.876 & 20.98 & 94.2 & 1.175 & 9.58 & 44.2 & 3.606 & 14.85 & 48.6 & 3.777 & 31.26 & 139.8 & 1.112 \\
    FastVGGT & 87.03 & 11.4 & 2.899 & 17.55 & 65.5 & 1.785 & 9.73 & 42.3 & 5.826 & 12.10 & 47.3 & 5.741 & 30.58 & 108.5 & 1.763 \\
    $\pi^3$ & 70.25 & \textbf{9.3} & 2.189 & 43.15 & 33.9 & 1.234 & 11.31 & \textbf{31.9} & 5.319 & 76.52 & \textbf{14.3} & 5.526 & 24.79 & 77.7 & 1.100 \\
    \midrule
    \dejaview ($K=16$) & 83.21 & \textbf{9.3} & 1.295 & \textbf{65.97} & \textbf{26.8} & 0.733 & \textbf{13.93} & 34.9 & 3.111 & 79.48 & 16.6 & 3.233 & \textbf{43.41} & 67.2 & 0.647 \\
    \dejaview ($K=14$) & 82.54 & \textbf{9.3} & 1.169 & 64.38 & 27.7 & 0.665 & 13.60 & 35.1 & 2.798 & \textbf{79.67} & 16.7 & 2.890 & 43.19 & \textbf{66.8} & 0.588 \\
    \midrule
    \ours, quality ($K=16$) & 84.58 & \textbf{9.3} & 1.016 & 62.53 & 29.8 & 0.586 & 13.77 & 34.8 & 2.352 & 78.06 & 17.1 & 2.420 & 42.55 & 67.6 & 0.524 \\
    \ours, balanced ($K=14$) & 83.42 & 9.5 & 0.843 & 60.36 & 31.6 & 0.494 & 13.39 & 35.1 & 1.853 & 77.43 & 17.4 & 1.894 & 42.10 & 68.1 & 0.448 \\
    \ours, fast ($K=14$) & 80.27 & 9.8 & \textbf{0.631} & 49.98 & 37.2 & \textbf{0.383} & 12.83 & 35.6 & \textbf{1.273} & 71.19 & 18.8 & \textbf{1.293} & 40.84 & 71.7 & \textbf{0.354} \\
    \bottomrule
  \end{tabular*}
\end{table*}

\Cref{tab:main} reports the three operating points of \ours. The balanced tier
runs $1.63\times$ faster than the 16-step model and lowers the average
AUC@$3^\circ$ from 57.20 to 55.34. The quality and fast tiers reach
56.30 and 51.02 at $1.31\times$ and $2.29\times$. The change differs
by dataset: DTU moves from 83.21 to 83.42 and ETH3D from 65.97 to
60.36. The speedup comes from cheaper updates: at the same 14 steps, the balanced tier is
$1.31$--$1.53\times$ faster than the original model and the fast tier
$1.66$--$2.24\times$.

\begin{figure}[t]
  \centering
  \includegraphics[width=0.70\columnwidth]{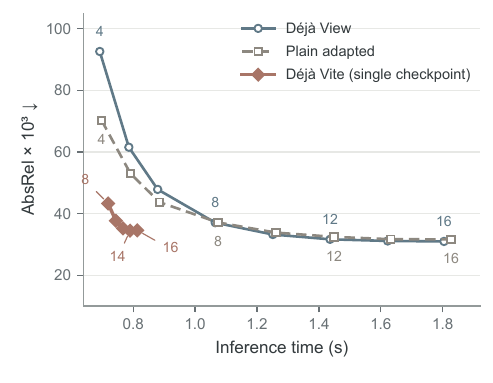}
  \caption{Geometry error versus full-model inference time across
  iteration budgets, with equal weighting of the five datasets.
  Numbers beside selected markers indicate $K$; all measured budgets
  are plotted. Each curve varies the budget of the corresponding
  model; the \ours curve uses a single adapted checkpoint. The corresponding
  pose curve appears in \cref{fig:teaser}(b).
  The original and plain models include $K=4,5,6$ to cover the
  latency range of the accelerated model.}
  \label{fig:pareto}
\end{figure}

\begin{table*}[t]
  \centering
  \footnotesize
  \setlength{\tabcolsep}{5pt}
  \renewcommand{\arraystretch}{0.90}
  \caption{Computational efficiency across input sizes. The standard
  benchmark reports the five-dataset mean with dataset-specific view
  counts; long-sequence results use 300 views for ScanNet++ and
  167--334 views for 7-Scenes. FLOPs are in tera-operations and peak
  GPU memory is in GiB. Parameters count the compact stored model;
  runtime memory includes the materialized weights used for inference.
  Peak memory is measured with eight-frame decoder chunks for all
  methods.}
  \label{tab:compute_efficiency}
  \begin{tabular*}{\textwidth}{@{\extracolsep{\fill}}lrrrrrrr@{}}
    \toprule
    Method & Params & \multicolumn{2}{c}{Standard benchmark}
      & \multicolumn{2}{c}{ScanNet++: 300 views}
      & \multicolumn{2}{c}{7-Scenes: 167--334 views} \\
    \cmidrule(lr){3-4}\cmidrule(lr){5-6}\cmidrule(lr){7-8}
    & (M) & FLOPs (T)$\downarrow$ & Peak (GiB)$\downarrow$
      & FLOPs (T)$\downarrow$ & Peak (GiB)$\downarrow$
      & FLOPs (T)$\downarrow$ & Peak (GiB)$\downarrow$ \\
    \midrule
    VGGT & 1190.6 & 572.3 & 8.94 & 8331 & 23.5 & 11699 & 26.9 \\
    AVGGT & 1190.6 & 220.8 & 9.08 & 1996 & 24.6 & 2606 & 28.1 \\
    FastVGGT & 1190.6 & 223.5 & 8.29 & 2192 & 22.7 & 3071 & 26.0 \\
    \midrule
    \dejaview ($K=16$) & 117.1 & 235.1 & 3.01 & 3564 & 9.2 & 5065 & 10.8 \\
    \ours, quality & 120.6 & 169.3 & 3.17 & 2430 & 10.75 & 3448 & 12.62 \\
    \ours, balanced & 120.6 & 125.1 & 3.17 & 1685 & 10.75 & 2387 & 12.62 \\
    \ours, fast & 120.6 & 76.2 & 3.17 & 879 & 10.75 & 1241 & 12.62 \\
    \bottomrule
  \end{tabular*}
\end{table*}

\noindent\textbf{Comparison at matched latency.}
The pose curve in \cref{fig:teaser}(b) and the geometry curve in
\cref{fig:pareto} compare reconstruction quality at each latency. We
interpolate the curves of the original model and of the plain baseline
at the latency of each configuration (\cref{tab:matched_latency}). The
balanced tier exceeds the plain curve by 4.7 points of AUC@$3^\circ$
and the original curve by 9.1, and it improves AUC@$30^\circ$, inlier
ratio, and AbsRel. A measured pair agrees: the balanced tier reaches
55.3 at 1.106 seconds, and the plain baseline at $K=10$ reaches 54.1
at 1.261 seconds. The quality tier gains 1.1 and 2.0 points. With one
adapted checkpoint swept over $K$ from 8 to 14, AUC@$3^\circ$ rises
from 42.35 to 52.00 while latency rises from 0.719 to 0.790 seconds
(\cref{tab:budget_curve}),
and all five budgets beat both interpolated baselines on all four
metrics.

\subsection{Ablations}
\label{sec:schedule_ablation}
\label{sec:components}
\noindent\textbf{Placement.}
We train schedules that differ only in where the approximate
operations sit, with the same training setup (\cref{tab:schedules}).
At 14 steps our schedule improves AUC@$3^\circ$ by 2.62 points over
uniform placement, 4.90 over reversed placement, 1.53 over a
stability-based rule, and 1.96 over a random schedule selected on
validation data, the best of eight random layouts by distillation
loss on held-out training scenes after 2,000 adaptation steps. Across 10, 12, 14, and 16 steps, all 16
comparisons have 95\% confidence intervals above zero
(\cref{app:schedule_ablation}). Latency differs from the uniform,
reversed, and random controls by at most 0.012 seconds. The placement findings of \cref{sec:analysis}
therefore hold after adaptation.

\begin{table}[t]
  \centering
  \footnotesize
  \setlength{\tabcolsep}{2pt}
  \renewcommand{\arraystretch}{0.90}
  \caption{Ablation of the three acceleration components of \ours at $K=14$.
  C: coarse-grid processing; P: KV pooling; R: attention-output reuse.
  Results weight the five datasets equally.
  AUC and inlier scores are percentages; AbsRel is scaled by $10^3$.
  Bold marks the best value in each metric.}
  \label{tab:component_ablation}
  \begin{tabular*}{\columnwidth}{@{\extracolsep{\fill}}cccrrrrr@{}}
    \toprule
    \multicolumn{3}{c}{Components} & \multicolumn{2}{c}{Pose AUC$\uparrow$}
    & Inlier$\uparrow$ & AbsRel$\downarrow$ & Time$\downarrow$ \\
    \cmidrule(lr){1-3}\cmidrule(lr){4-5}
    C & P & R & @$3^\circ$ & @$30^\circ$ & (\%) & $\times10^3$ & (s) \\
    \midrule
    \NA & \NA & \NA & \textbf{56.71} & 91.64 & 79.10 & 31.68 & 1.629 \\
    \midrule
    \mbox{\normalsize$\checkmark$} & \NA & \NA & 54.24 & 91.15 & 77.74 & 32.60 & 1.173 \\
    \NA & \mbox{\normalsize$\checkmark$} & \NA & 55.93 & 91.47 & 78.67 & 32.22 & 1.410 \\
    \NA & \NA & \mbox{\normalsize$\checkmark$} & 56.52 & \textbf{91.65} & \textbf{79.33} & \textbf{31.36} & 1.497 \\
    \midrule
    \mbox{\normalsize$\checkmark$} & \mbox{\normalsize$\checkmark$} & \NA & 53.58 & 90.88 & 76.98 & 33.46 & 0.932 \\
    \mbox{\normalsize$\checkmark$} & \NA & \mbox{\normalsize$\checkmark$} & 53.20 & 90.95 & 77.02 & 33.16 & 1.029 \\
    \NA & \mbox{\normalsize$\checkmark$} & \mbox{\normalsize$\checkmark$} & 55.26 & 90.76 & 76.74 & 35.22 & 1.266 \\
    \midrule
    \mbox{\normalsize$\checkmark$} & \mbox{\normalsize$\checkmark$} & \mbox{\normalsize$\checkmark$} & 52.00 & 90.61 & 76.04 & 34.46 & \textbf{0.790} \\
    \bottomrule
  \end{tabular*}
\end{table}

\noindent\textbf{Components.}
\Cref{tab:component_ablation} evaluates all combinations of the three
operations at 14 steps. Reuse is nearly free: it lowers AUC@$3^\circ$
by 0.19 points and latency from 1.629 to 1.497 seconds. Coarse-grid
processing costs 2.47 points and brings latency to 1.173 seconds. The
three together reach 0.790 seconds. The operations interact in
geometry: on ETH3D, pooling with reuse gives an AbsRel of 0.0449,
against 0.0305 and 0.0291 for each alone (\cref{app:components}).

\subsection{Comparison and Limitations}
\label{sec:external_comparison}
\label{sec:limits}

In the dataset average, the balanced tier has higher pose AUC and
lower latency than VGGT, FastVGGT, AVGGT, and $\pi^3$
(\cref{tab:external_macro}). Individual datasets differ. VGGT reaches
96.48 on DTU against our 83.42, and $\pi^3$ has better geometry on
ScanNet++ (\cref{tab:external_dataset}). On 300-view ScanNet++ inputs, the quality, balanced, and
fast tiers reduce FLOPs by 31.8\%, 52.7\%, and 75.3\%
(\cref{tab:compute_efficiency}). Peak memory rises from 9.2 to 10.75
GiB for the balanced tier. The multistep history, attention cache, and
adapter execution contribute to the memory footprint
(\cref{app:limits}).

The accuracy cost is concentrated. Against the 16-step model, the
balanced tier wins, ties, and loses on 41, 13, and 99 of the 153
scenes, and it loses on 12 of the 13 ETH3D scenes. Scenes with low
covisibility lose more, 2.50 points against 1.15 for high
covisibility (\cref{app:limits}).

\section{Conclusion}
\label{sec:conclusion}
Our interventions show that the steps of \dejaview are interchangeable
refinements and that later steps absorb approximation errors. The loop
shrinks these errors in all 12 tested conditions, and more than random
perturbations of equal magnitude, because its channel gates scale down
the channels that carry them. At the output, the direction of the
remaining error sets its cost: at equal norm it changes the predicted
rotations 1.55 times as much as a random error. Spatial approximations
are cheapest early in the loop, and reuse is cheapest when its steps
are spaced apart. Guided by these findings, \ours combines coarse-grid
processing, key--value pooling, and attention-output reuse with
lightweight self-distillation. Across five benchmarks, the balanced
configuration runs $1.63\times$ faster than the 16-step model with a
1.86-point reduction in average AUC@$3^\circ$, and at equal latency it
exceeds a trained full-block baseline by 4.73 points.

\clearpage %
\section*{LLM Usage Statement}
We used Opus 5.5 and GPT-6 Astra to assist with drafting pseudocode
and polishing the manuscript's language, including grammar and clarity.
The authors reviewed and verified all AI-assisted content and take full
responsibility for the accuracy, originality, and integrity of the work.

{\small
  \bibliographystyle{ieeenat_fullname}
  \bibliography{references}
}

\setcounter{figure}{0}\setcounter{table}{0}
\renewcommand{\thefigure}{S\arabic{figure}}\renewcommand{\thetable}{S\arabic{table}}
\renewcommand{\theHfigure}{S\arabic{figure}}\renewcommand{\theHtable}{S\arabic{table}}
\clearpage
\appendix
\renewcommand{\dbltopfraction}{0.66}
\renewcommand{\dblfloatpagefraction}{0.66}
\renewcommand{\topfraction}{0.66}
\renewcommand{\floatpagefraction}{0.66}

\section{Interventions in the Reconstruction Loop}
\label{app:mechanism}

We follow each approximation from the step where it enters the loop to the pose and geometry decoded at the output. The paired runs in \cref{app:protocol} provide a common reference for the single-step and schedule comparisons.

\subsection{Paired trajectories and error measurements}
\label{app:protocol}

We compare the released checkpoints of \dejaview, VGGT~\cite{wang2025vggt},
$\pi^3$~\cite{wang2025pi3}, and DA3~\cite{lin2025da3} through paired
interventions.

\noindent\textbf{Analysis data.}
Our analysis uses 65 scenes, with 13 each from DTU, ETH3D, 7-Scenes,
ScanNet++, and nuScenes, with at most 32 input views per scene; scenes
with fewer views use all of them. We use $K=16$ for \dejaview. Within each dataset, even-indexed scenes are used to
derive placement rules and odd-indexed scenes to evaluate their
predictions. Comparisons are paired by scene, and uncertainty in output
deviations is summarized with hierarchical bootstrap confidence intervals.

\noindent\textbf{Paired interventions.}
For each scene, we run an unmodified reference trajectory and a perturbed
trajectory with identical images and view ordering. At a selected step
$u$, we replace one operation and then resume the original computation
for the remaining steps. Interventions include token pooling, restricting
cross-view attention, and skipping an update. In \dejaview, we additionally
probe cross-step reuse of keys, values, attention probabilities, and
attention outputs. We vary the intervention position while keeping its
type and strength fixed. Multiple-intervention experiments further vary
the placement and spacing of a fixed number of approximate steps.
KV subsampling follows the second step of AVGGT~\cite{sun2026avggt}:
keys and values are subsampled on a $2\times2$ grid, the first frame
and the diagonal of the attention matrix are kept, and one mean
key--value token is added.

\noindent\textbf{Local error model.}
Let $f_t$ denote the update with its time inputs fixed and let
$\delta_t=\widetilde z_t-z_t$ compare perturbed and clean trajectories.
After a single approximation at step $u$, subsequent exact updates give
\begin{equation}
  \delta_{t+1}=J_t\delta_t+\mathcal R_t(\delta_t),\qquad
  J_t=Df_t(z_t),\quad t>u,
  \label{eq:local_step}
\end{equation}
where $\mathcal R_t(\delta)=o(\|\delta\|_2)$ at a differentiable
reference state. Norms in this local analysis are Euclidean norms of
vectorized feature tensors. Defining the downstream Jacobian product
$\Phi_{K,u+1}=J_{K-1}\cdots J_{u+1}$ yields
\begin{equation}
  \delta_K\approx\Phi_{K,u+1}\delta_{u+1},\qquad
  \gamma_t(e)=\frac{\|J_te\|_2}{\|e\|_2}.
  \label{eq:local_propagation}
\end{equation}
The directional gain $\gamma_t(e)$ measures attenuation or amplification
along the injected direction. Error magnitude and direction therefore
both influence what reaches the final state.

\noindent\textbf{Error measurements.}
Let $\widetilde z_t$ denote the perturbed state and
$\delta_t=\widetilde z_t-z_t$ its deviation from the reference at the
corresponding boundary. For the set $\mathcal P$ of image-patch tokens,
we compute the mean relative deviation and reference-token norm:
\begin{equation}
  R_t=\frac{1}{|\mathcal P|}\sum_{i\in\mathcal P}
      \frac{\|\delta_{t,i}\|_2}{\|z_{t,i}\|_2},
  \quad
  m_t=\frac{1}{|\mathcal P|}\sum_{i\in\mathcal P}\|z_{t,i}\|_2.
  \label{eq:state_measurements}
\end{equation}
The scale-adjusted deviation proxy is $\widehat A_t=R_t m_t$.
It combines relative token error with the mean reference-state scale.
We track its stepwise gain as
\begin{equation}
  g_t=\frac{\widehat A_{t+1}}{\widehat A_t}.
  \label{eq:perturbation_gain}
\end{equation}
Within each scene, step gains are summarized by their geometric mean
and the fraction below one. Scene-level summaries use medians; pooled
summaries combine all valid scene--step records. We exclude missing
and nonfinite values, and omit step ratios with a zero numerator or
denominator. At the output, we measure rotation deviations from the
reference predictions and changes in pose AUC@$3^\circ$,
AUC@$30^\circ$, and AbsRel against ground truth.

\subsection{What changes when one update is removed?}
\label{app:loop_properties}

We compare the most sensitive update in each scene, then untie the loop weights to test whether late updates acquire a specialized role.

This section gives the protocols behind \cref{sec:loop_properties}.
The skip, cross-view ablation, and stationarity measurements use the
released checkpoints on the 65 analysis scenes with at most 32 views. A
\emph{unit} is one loop step in \dejaview and one pair of a
frame-attention block and a global-attention block in the stacked
models. Each intervention is applied to a single unit, and every other
unit runs exactly.

\noindent\textbf{Skip.}
Skipping a unit leaves the state unchanged at that unit. In \dejaview
the loop index still advances, so the following steps receive their
original time inputs. For each scene we skip every unit in turn, record
the drop in AUC@$30^\circ$ against ground truth, and keep the largest
drop over units. \Cref{tab:loop_properties_full} reports the median of
this worst-unit drop over scenes, its maximum, and the number of scenes
in which it exceeds 5 points. The largest worst-unit drop of \dejaview
over all 65 scenes is 1.32 points, and every scene of every stacked
model has a unit whose skip costs more than that.

\noindent\textbf{Cross-view ablation.}
This intervention removes what the other views contribute at one unit
and keeps the activation statistics. For the queries of a view, the
keys and values of all other views are replaced by those of a donor
scene recorded at the same unit, and the keys and values of the query's
own view are kept. The donor is another scene of the same dataset with
the same token grid and at least as many views when one is available.
Using a scene as its own donor reproduces exact attention with zero
error, which verifies the implementation. The statistic is the same
worst-unit drop in AUC@$30^\circ$ as for the skip.

\noindent\textbf{Attention stationarity.}
On the clean pass we record the global-attention rows of 128 query
patches with ground-truth correspondences at every unit, and compute
the cosine similarity between the rows of consecutive units. The onset
$k^\star$ of a scene is the first unit from which every later
similarity is at least 0.95. We call the attention pattern stationary
when this holds over at least the last three units. In \dejaview all
65 scenes are stationary, with a median onset at relative depth
$k^\star/(U-1)=0.27$, where $U$ is the number of units. None of the 195
scene--model pairs of the stacked models is stationary. The update
statistics come from a 16-scene subset, four scenes from each of four
datasets. From the first step to the last, the relative update size
$\|z_{t+1}-z_t\|/\|z_t\|$ of \dejaview falls from 0.76 to 0.07, and the
cosine between consecutive updates rises from 0.17 to 0.98. In the
stacked models the update size stays between 0.24 and 1.8 without
decay, and consecutive updates of VGGT and $\pi^3$ are close to
uncorrelated, with cosines between $-0.3$ and $0.3$.

\begin{table}[t]
  \centering
  \small
  \setlength{\tabcolsep}{3pt}
  \caption{Worst-unit interventions and attention stationarity on 65
  scenes. Median and maximum are over scenes of the largest AUC@$30^\circ$
  drop over units, in points. $>5$ counts scenes whose worst-unit drop
  exceeds 5 points.}
  \label{tab:loop_properties_full}
  \begin{tabular}{lrrrrr}
    \toprule
    & \multicolumn{2}{c}{Skip} & \multicolumn{2}{c}{Cross-view} & Stationary \\
    \cmidrule(lr){2-3}\cmidrule(lr){4-5}
    Model & Median & $>5$ & Median & $>5$ & scenes \\
    \midrule
    \dejaview & 0.18 & 0/65 & 1.9 & 3/65 & 65/65 \\
    VGGT & 48.8 & 65/65 & 15.9 & 50/65 & 0/65 \\
    $\pi^3$ & 54.8 & 65/65 & 11.9 & 56/65 & 0/65 \\
    DA3 & 82.2 & 65/65 & 7.5 & 38/65 & 0/65 \\
    \bottomrule
  \end{tabular}
\end{table}

\noindent\textbf{Tied and untied weights.}
All variants start from the released \dejaview block and run eight
steps. The tied model shares one block across the eight steps. The
untied model holds eight copies of the block, one per step, which are
identical at initialization and give the same output as the tied model
before training. Both are trained by distillation from the released
model with the same data, loss, and schedule: 20,000 steps at a
learning rate of $10^{-4}$, then 60,000 steps at $5\times10^{-5}$. We
train the untied model with two seeds, and evaluate on 32 scenes.

\begin{table}[t]
  \centering
  \small
  \setlength{\tabcolsep}{3pt}
  \caption{AUC@$3^\circ$ drop, in points, from skipping one step of
  eight-step models fine-tuned from the \dejaview block for 80,000
  steps, on 32 scenes. The last column is the AUC@$3^\circ$ of the
  unmodified model.}
  \label{tab:tied_untied}
  \begin{tabular}{lrrrrrrrrr}
    \toprule
    Skipped step & 1 & 2 & 3 & 4 & 5 & 6 & 7 & 8 & AUC \\
    \midrule
    Tied & 4 & 3 & 3 & 3 & 3 & 3 & 4 & 4 & 53.2 \\
    Tied, no time & 4 & 4 & 4 & 4 & 4 & 4 & 4 & 4 & 52.8 \\
    Untied, seed 0 & 7 & 4 & 4 & 4 & 5 & 6 & 8 & 19 & 46.2 \\
    Untied, seed 1 & 9 & 5 & 5 & 5 & 5 & 7 & 8 & 18 & 47.7 \\
    \bottomrule
  \end{tabular}
\end{table}

\Cref{tab:tied_untied} reports the AUC@$3^\circ$ drop from skipping
each step after 80,000 training steps. The tied model stays flat at 3
to 4 points. The untied model develops a specialized last block, whose
skip costs 19 and 18 points in the two seeds. This specialization
grows with training: the last-block cost is 15 and 12 points after
20,000 steps, 17 and 17 after 50,000, and 19 and 18 after 80,000.
Swapping the last two blocks of the untied model costs 12 points and
repeating its last block costs 10 to 11 points, while both operations
change the tied model by at most 1 point. The updates of the untied
last block also lose their alignment with the preceding update, with a
cosine of $-0.05$ and $-0.02$ against 0.78 for the tied model. The
untied model has eight times the block parameters and is less
accurate, with AUC@$3^\circ$ of 46.2 and 47.7 against 53.2.

\noindent\textbf{Time conditioning and partial weight sharing.}
Two further controls isolate weight sharing. A tied model trained
without time conditioning reaches 52.8 and loses 4 points for every
skipped step. A tied model trained with a random number of steps, as
in the original recipe, loses 2 to 3 points per skipped step after
20,000 training steps, against 3 to 5 for a fixed number of steps.
Giving only the last step its own block reproduces the specialization
of the untied model in that block, with a skip cost of 17 points in
both seeds, while the seven shared steps keep a flat cost of 5 to 7
points. The interchangeable steps therefore follow from weight
sharing, and a free block becomes a specialized output stage.

\subsection{Propagation after a single approximation}
\label{app:propagation}

\begin{figure*}[!htbp]
  \centering
  \includegraphics[width=\textwidth]{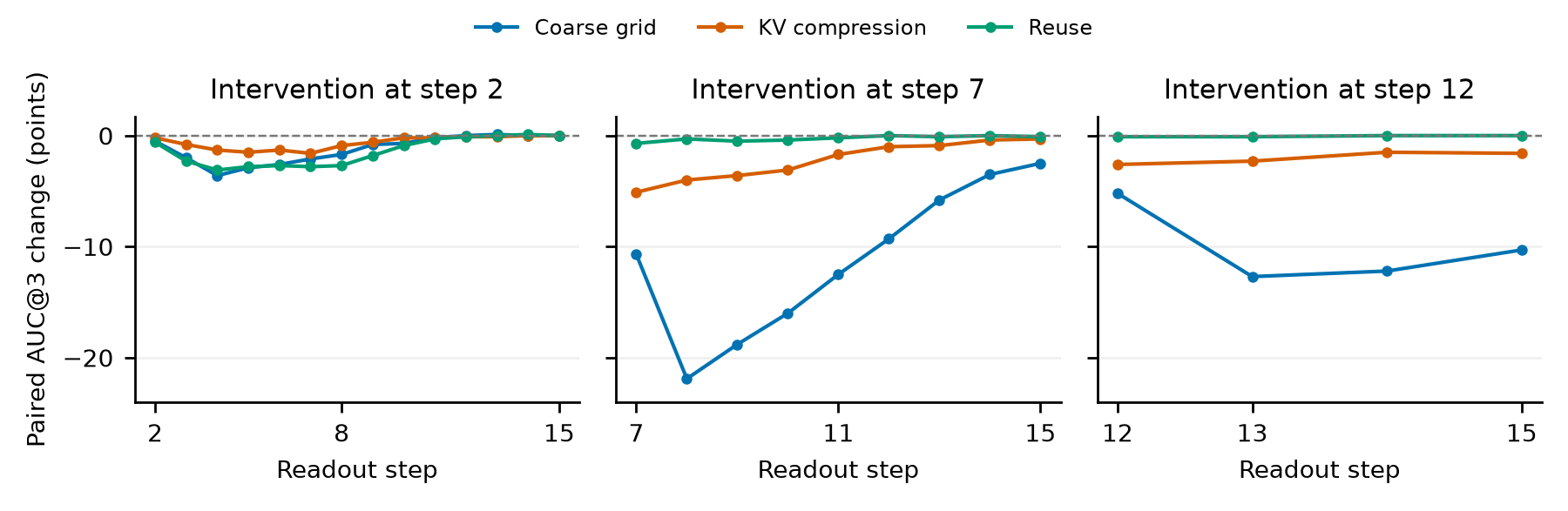}
  \caption{FP32 task recovery in the original 16-step model. Each panel
  applies one intervention at the indicated step. Curves show the
  scene median of the paired AUC@$3^\circ$ difference against ground
  truth, in percentage points, over 65 scenes with at most 32 views each.
  Zero indicates no difference at the reported precision.}
  \label{fig:task_recovery}
\end{figure*}

\Cref{tab:matched_perturbations} measures the fraction of an injected state error that reaches the final step. \Cref{fig:task_recovery,tab:task_recovery} track the corresponding task-level changes. After a coarse-grid update at iteration 7, the median AUC@$3^\circ$ change is $-21.9$ points one step later and $-2.5$ at the output. Moving the update to iteration 12 leaves a final loss of 10.3 points.

\noindent\textbf{Matched-magnitude perturbations.}
We inject errors produced by approximate operations at early, middle,
and late positions, and compare them with random perturbations of the
same magnitude. Interventions are applied at steps 2, 7, and 12 of
the 16-step \dejaview model, and at 15\%, 50\%, and 85\% of depth
in the stacked models. All propagation measurements in this experiment
use FP32. For each scene, we divide the final scale-adjusted deviation
$\widehat A$ by its value immediately after injection.
\Cref{tab:matched_perturbations} reports the median of these
scene-level ratios. Scenes with an injected deviation at or below
$10^{-9}$ are excluded. Each matched-magnitude condition uses one fixed random direction. Comparisons between structured and random
conditions use their respective scene medians.

At the operators' native perturbation magnitude, \dejaview reduces
the structured-error proxy in all 12 operator--position conditions, with
retained fractions of 0.29--0.85. Each is smaller than its matched
random control. Across four perturbation magnitudes and the same
conditions, the structured error has a smaller retained fraction in
47 of 48 comparisons. The exception is doubled KV-pooling error
at step 12, with fractions of 1.25 for the structured perturbation
and 1.14 for the random control.

The stacked models increase the scale-adjusted structured-error proxy
in 17 of 18 native-magnitude conditions. The exception is late KV subsampling
in VGGT, with a retained fraction of 0.96. Reference-state growth can
make these deviations appear smaller under state-relative measures:
in VGGT, the reference norm grows by factors of 24.8, 14.5, and 3.6
from the early, middle, and late injection boundaries to the output.
The relative deviation and its scale-adjusted proxy therefore
characterize different aspects of the trajectory. Directional preferences also vary:
VGGT retains more structured than random error in five of six
conditions, DA3 in all six, and $\pi^3$ exhibits mixed responses.

\begin{table*}[!htbp]
  \centering
  \small
  \setlength{\tabcolsep}{5pt}
  \caption{FP32 retained scale-adjusted deviation $\widehat A$ at
  the native perturbation magnitude. Each entry is the median of
  per-scene final-to-injected ratios over the valid scenes.
  E/M/L denote steps 2/7/12 in \dejaview and 15/50/85\% of depth in
  stacked models. Random controls match the injected error magnitude.
  Values below one indicate attenuation.}
  \label{tab:matched_perturbations}
  \begin{tabular}{llrrrrrr}
    \toprule
    Model & Approximation & \multicolumn{3}{c}{Structured error}
    & \multicolumn{3}{c}{Random control} \\
    \cmidrule(lr){3-5}\cmidrule(lr){6-8}
    & & E & M & L & E & M & L \\
    \midrule
    \dejaview & KV pooling & 0.37 & 0.67 & 0.85 & 0.43 & 0.94 & 1.05 \\
     & KV subsampling & 0.39 & 0.53 & 0.68 & 0.43 & 0.98 & 1.11 \\
     & Coarse grid & 0.29 & 0.55 & 0.72 & 1.02 & 1.52 & 1.16 \\
     & Reuse & 0.30 & 0.42 & 0.62 & 0.41 & 0.94 & 1.02 \\
    \midrule
    VGGT & KV pooling & 6.00 & 6.34 & 1.38 & 3.18 & 2.14 & 1.10 \\
     & KV subsampling & 3.94 & 4.39 & 0.96 & 3.20 & 2.02 & 1.10 \\
    \midrule
    $\pi^3$ & KV pooling & 2.54 & 1.16 & 1.80 & 1.46 & 1.55 & 2.19 \\
     & KV subsampling & 1.37 & 1.56 & 1.65 & 1.47 & 1.56 & 1.90 \\
    \midrule
    DA3 & KV pooling & 15.80 & 6.58 & 3.05 & 7.22 & 2.76 & 1.23 \\
     & KV subsampling & 13.80 & 6.63 & 2.72 & 7.43 & 2.81 & 1.24 \\
    \bottomrule
  \end{tabular}
\end{table*}

\begin{table*}[!htbp]
  \centering
  \small
  \caption{Final task changes and recovery after a single intervention
  in the original model. Task changes are medians of paired scene
  differences. Rotation and AUC@$30^\circ$ recovery ratios compare the
  final output with the first post-intervention readout, using the
  eligibility rules in the text. The final column counts rotation
  recovery without task recovery out of all 65 scenes.}
  \label{tab:task_recovery}
  \begin{tabular}{lrrrrr}
    \toprule
    Intervention & $\Delta$AUC@$30^\circ$ (pt) & $\Delta$AbsRel $\times10^3$ & Rotation recovery & Task recovery & Count \\
    \midrule
    Coarse@2 & $+0.02$ & $-0.00$ & $0.03$ & $-0.00$ & 2/65 \\
    Coarse@7 & $-0.40$ & $+1.55$ & $0.19$ & $0.08$ & 0/65 \\
    Coarse@12 & $-1.40$ & $+1.51$ & $0.50$ & $0.56$ & 15/65 \\
    KV pooling@2 & $0.00$ & $+0.02$ & $0.03$ & $0.00$ & 1/65 \\
    KV pooling@7 & $-0.05$ & $+0.26$ & $0.18$ & $0.07$ & 5/65 \\
    KV pooling@12 & $-0.26$ & $+0.42$ & $0.54$ & $0.31$ & 12/65 \\
    Reuse@2 & $+0.01$ & $+0.04$ & $0.02$ & $-0.00$ & 0/65 \\
    Reuse@7 & $-0.01$ & $+0.01$ & $0.14$ & $0.00$ & 1/65 \\
    Reuse@12 & $-0.01$ & $-0.00$ & $0.55$ & $0.08$ & 3/65 \\
    \bottomrule
  \end{tabular}
\end{table*}

\subsection{Channel gating and the spatial structure of error}
\label{app:gates}

Pooling and random perturbations of equal norm distribute their energy differently across channels. We compare their responses along the gate path, then shuffle pooling residuals across tokens to preserve channel energy while changing spatial arrangement.

\noindent\textbf{Directions and propagation.}
We evaluate Jacobian--vector products in the released $K=16$ model
using FP32 on the same 65 scenes with at most 32 views each. Pooling errors
are injected after step 7 or 12, leaving eight or three updates,
respectively. We propagate the injected direction through Jacobians
evaluated along the clean trajectory and compare it with finite-error
propagation. Four random directions match the injected error norm of
each token. A shuffled control permutes the real error's token vectors
within each view, preserving its per-channel energy. An additional
control draws directions within the subspace containing 90\% of the
real error's energy.

\noindent\textbf{Gate and branch decomposition.}
Let $G_t$ be the diagonal operator formed by the product of the two
channel gates of the block (\cref{sec:looped_model}), repeated across
token positions. The local response decomposes as
\begin{equation}
  J_te=G_te+b_t(e),\qquad b_t(e)=(J_t-G_t)e.
  \label{eq:gate_branch_split}
\end{equation}
The term $b_t$ includes the attention and feed-forward responses and
their interactions through the two subblocks. Its effect on squared
error norm follows directly:
\begin{equation}
  \|J_te\|_2^2-\|G_te\|_2^2
  =2\langle G_te,b_t(e)\rangle+\|b_t(e)\|_2^2.
  \label{eq:gate_branch_energy}
\end{equation}
A response opposing the incoming error can introduce energy in other
directions. Net attenuation relative to the gate path depends on both
terms in \cref{eq:gate_branch_energy}.

To compare complete trajectories, let $L(e)$ be the gain under the
Jacobian product in \cref{eq:local_propagation}, and let $M(e)$ be
the gain under the corresponding product of gate matrices. For real
and control directions, their gain ratio factors scene by scene as
\begin{equation}
  \frac{L(e_{\rm real})}{L(e_{\rm ctrl})}
  =\frac{M(e_{\rm real})}{M(e_{\rm ctrl})}
   \frac{L(e_{\rm real})/M(e_{\rm real})}
        {L(e_{\rm ctrl})/M(e_{\rm ctrl})}.
  \label{eq:gate_relative_factor}
\end{equation}
The second factor measures the full-trajectory response relative to
its gate path, including branch interactions. We compute factors
within scenes before summarizing them; products of reported medians
need not equal the median total ratio.

\noindent\textbf{Channel distribution and spatial arrangement.}
For step-12 injection, the median real and random linear gains are
0.905 and 1.195; their gate-path gains are 0.469 and 0.767.
The median scene-wise real-to-random ratio is 0.758, with a gate
factor of 0.611 and a gate-relative factor of 1.232. The gate factor
is below one and the gate-relative factor above one in all 65 scenes.
At step 7, the corresponding factors are 0.296 and 1.940, again with
the same ordering in every scene. The gate path supplies the
real-to-random attenuation advantage, while the remaining response
offsets part of that advantage.

\begin{figure*}[!htbp]
  \centering
  \includegraphics[width=\textwidth]{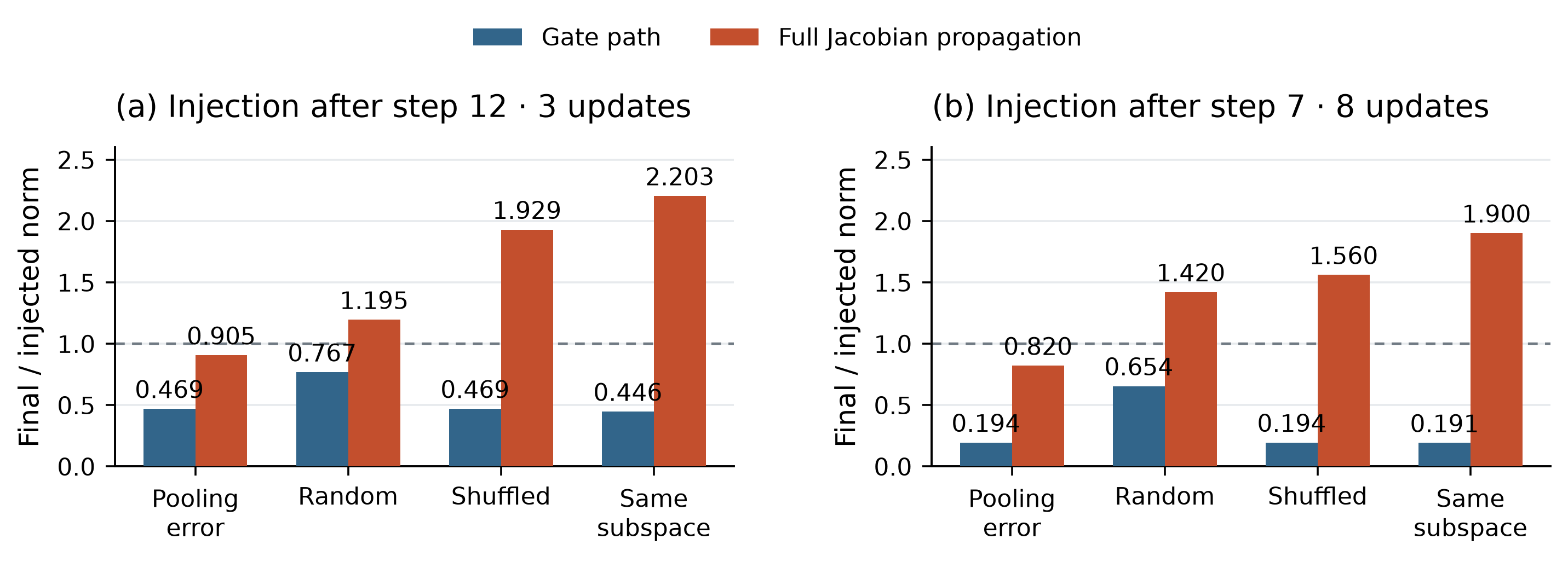}
  \caption{Gate-path and full Jacobian propagation of pooling errors and
  control directions. Panels show injections after steps 12 and 7 in
  the original $K=16$ model, using FP32 on 65 scenes with at most 32 views.
  Bars report median final-to-injected norm gains along the clean
  reference trajectory; the dashed line marks unit gain. Shuffling
  preserves per-channel energy and the gate-path gain, while changing
  the full propagation response.}
  \label{fig:propagation}
\end{figure*}

For real versus shuffled directions, the gate factor is one. The
median total gain ratio is 0.468 after step-12 injection and 0.534
after step-7 injection, with the real direction smaller in 65 and
64 scenes, respectively. For the same-subspace control, gate factors
are 1.050 and 1.015, while gate-relative factors are 0.394 and 0.428.
The branch response therefore depends on the spatial organization
of the error even when channel energy is preserved or approximately
matched. The reported first-step branch-response cosine with the
real error is approximately $-0.54$: directional opposition coexists
with a larger total norm in the gate-path comparison (\cref{app:extra_gates}).
Channel energy distribution governs the gating contrast, while the
remaining response distinguishes spatial arrangements
(\cref{fig:propagation}).

\subsection{From residual direction to decoded geometry}
\label{app:output}

The prediction heads read the residual left by the loop. We hold the final residual norm fixed and change its direction before decoding the same clean scene state.

Let $y=H(z_K)$ denote decoded geometry, with rotations represented
in local coordinates. Linearizing the prediction heads gives
\begin{equation}
  \Delta y\approx DH(z_K)\Phi_{K,u+1}\delta_{u+1}.
  \label{eq:decoded_error}
\end{equation}
For a unit final-state direction $v$, define its output sensitivity
as $s_H(v)=\|DH(z_K)v\|_2$. Then
\begin{equation}
  \|\Delta y\|_2\approx\|\delta_K\|_2\,
  s_H\!\left(\frac{\delta_K}{\|\delta_K\|_2}\right).
  \label{eq:output_direction_gain}
\end{equation}
Sensitivity is evaluated separately for each output quantity.

\noindent\textbf{Directional sensitivity of the prediction heads.}
To isolate the influence of the final error direction, we take residual
state deviations produced by structured perturbations and by four
random controls, match their total norms to the structured residual
at the final state, and decode with the same prediction heads.
A second normalization matches each token's norm separately; both
normalizations use a denominator floor of $10^{-12}$. Further
controls use two fresh isotropic random directions and one spatially
shuffled structured residual. Directions are added to the same clean
final state at scales $\alpha\in\{0.25,0.5,1,2\}$. For each
scene, random outputs are summarized by their median. Response ratios
are computed against this within-scene median and then summarized by
the median across scenes; counts use a strict greater-than comparison.

At $\alpha=1$, the structured residual produces a larger rotation
deviation than the propagated-random comparison in 62 of 65 scenes,
a larger relative-translation deviation in 62, and a larger depth
deviation in 58 (\cref{tab:head_direction}). Matching each token's
norm separately yields larger rotation deviation in 62 scenes and
larger translation deviation in 57. The directional sensitivity thus
persists under both whole-state and token-wise magnitude controls.
Together with the trajectory interventions, these measurements
identify three factors in approximation cost: the injected deviation,
its subsequent propagation, and the sensitivity of the decoded
outputs to the remaining error direction.
At $\alpha=1$ with total-norm matching, the median structured-to-random
response ratio is 1.55 for rotation, 1.81 for camera centers, 1.36
for pairwise rotation, 1.43 for pairwise translation, and 1.25 for
depth. All five comparisons have positive random-reference deviations
in every scene. Token-wise norm matching gives rotation and translation
ratios of 1.57 and 1.44, respectively.

\begin{table}[!htbp]
  \centering
  \small
  \caption{Final-state directional sensitivity at $\alpha=1$ with
  matched total residual norms. Ratios are medians of scene-wise
  structured responses divided by the median of four propagated-random
  responses. Counts use the same scenes and strict comparisons.
  Random-reference deviations are positive in all 65 scenes.}
  \label{tab:head_direction}
  \begin{tabular}{lrr}
    \toprule
    Output deviation & Ratio & Structured larger \\
    \midrule
    Rotation & 1.55 & 62/65 \\
    Camera center & 1.81 & 59/65 \\
    Pair rotation & 1.36 & 60/65 \\
    Pair translation & 1.43 & 62/65 \\
    Depth & 1.25 & 58/65 \\
    \bottomrule
  \end{tabular}
\end{table}

\subsection{Placing spatial approximations along the loop}
\label{app:placement}

\begin{figure*}[!htbp]
  \centering
  \includegraphics[width=\textwidth]{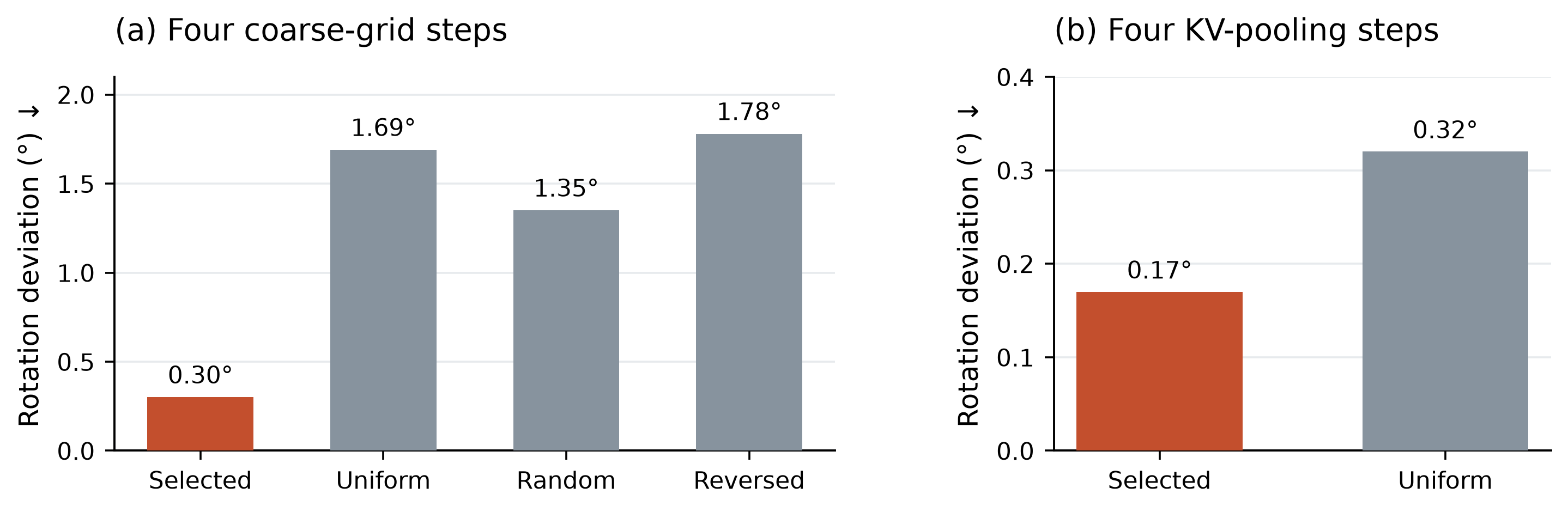}
  \caption{Placement of spatial approximations on 30 held-out scenes.
  Each panel fixes the number and type of approximate updates and
  reports output rotation deviation from the unmodified model.
  The selected layout outperforms uniform placement on all 30 scenes
  for coarse-grid processing and on 29 scenes for KV pooling.
  The two panels use different vertical scales.}
  \label{fig:placement}
\end{figure*}

We fix the operator and the number of approximate steps, move them along the trajectory, and rank layouts by the rotation deviation of the decoded poses (\cref{fig:placement}).

For an approximate update $\widetilde f_t$, define the additional
error evaluated at the perturbed state as
$\epsilon_t=\widetilde f_t(\widetilde z_t)-f_t(\widetilde z_t)$.
Equation~\eqref{eq:local_step} becomes
\begin{equation}
  \delta_{t+1}=J_t\delta_t+\epsilon_t+\mathcal R_t(\delta_t).
  \label{eq:multiple_injections}
\end{equation}
Thus, each approximation contributes an immediate deviation that
passes through its remaining updates. For cached computation,
$\epsilon_t$ also depends on cache history and on deviations already
present in the state.

For an intervention at step $u$, we summarize downstream propagation by
the retained fraction
\begin{equation}
  r_u=\frac{\widehat A_K}{\widehat A_{\mathrm{inj}}},
  \label{eq:retained_fraction}
\end{equation}
where $\widehat A_{\mathrm{inj}}$ is measured at the immediate
post-injection boundary.

We next evaluate whether these observations guide the placement of
approximate computation. With four coarse-grid steps, the selected
placement yields a rotation deviation of $0.30^\circ$ from the reference,
compared with $1.69^\circ$ for uniform placement, $1.35^\circ$ for random
placement, and $1.78^\circ$ for reversed placement. It outperforms
uniform placement on all 30 held-out scenes. For four KV-pooling steps,
the corresponding deviations are $0.17^\circ$ and $0.32^\circ$, with
improvement on 29 of 30 scenes.

The preferred position depends on the operator. Moving coarse-grid
processing from early to late steps increases output deviation by
$1.41^\circ$, with the early placement preferred in 64 of 65 scenes.
Moving reuse from late to early steps increases deviation by
$0.41^\circ$, with late placement preferred in 60 of 65 scenes.
Coarse computation benefits from subsequent recovery; late reuse
benefits from a smaller immediate disturbance.

The early-placement advantage varies with the iteration budget and
with the schedule already present in an adapted model.
\Cref{app:prediction,app:transfer} report cost prediction on held-out
operators, transfer across budgets and view counts, and long-sequence
evaluation.

\subsection{Spacing repeated attention reuse}
\label{app:reuse}

\begin{figure*}[!htbp]
  \centering
  \includegraphics[width=\textwidth]{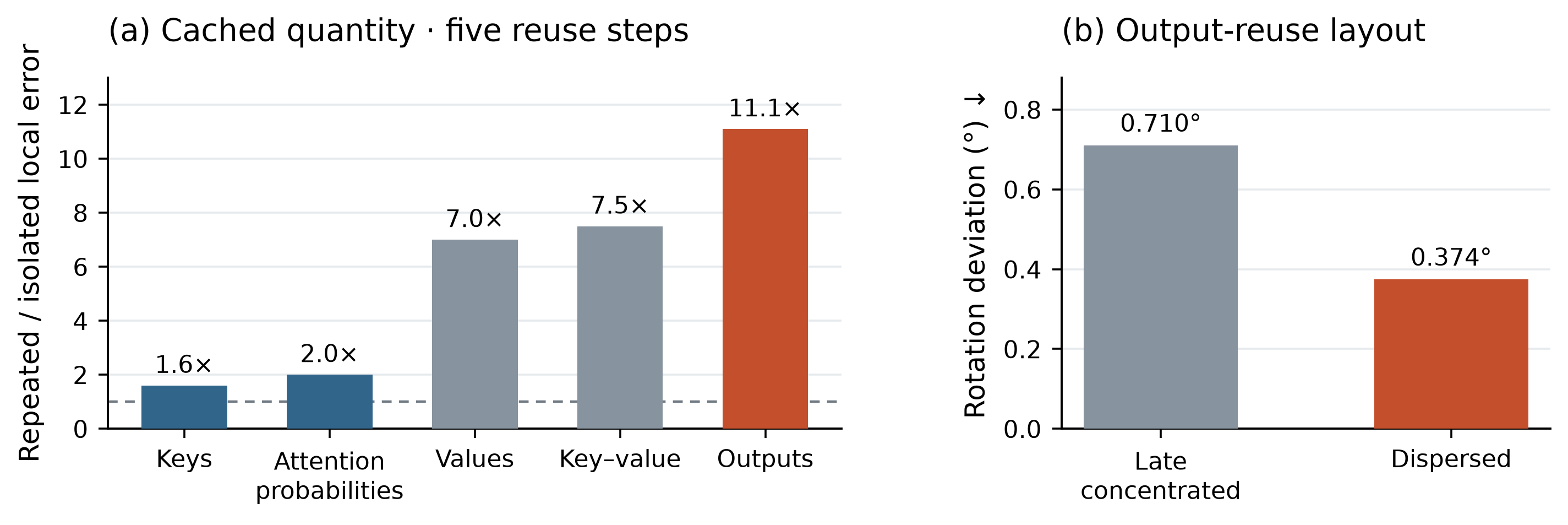}
  \caption{Composition of cross-step reuse errors. (a) Local-error
  amplification at the final reuse step in the layout
  $\{6,8,10,12,14\}$, relative to applying that intervention in
  isolation. The dashed line marks equal local error. (b) Output
  rotation deviation for two layouts with five output-reuse steps
  on 30 held-out scenes; dispersed reuse wins on 25 scenes.}
  \label{fig:cache}
\end{figure*}

We compare the error at the last reuse step with the error of an isolated reuse at that same step. This comparison measures the contribution of cache history. A second comparison fixes the number of reuse steps and measures the effect of spacing on the final poses (\cref{fig:cache}).

Cross-step reuse couples the current approximation to earlier states.
To characterize this dependence, we compare five cached quantities:
keys, values, key--value pairs, attention probabilities, and projected
attention outputs. Attention probabilities specify the weights applied
to values; caching them still allows the values to be recomputed from
the current state.

For the five-reuse layout at steps $\{6,8,10,12,14\}$, the local error
of output reuse at the final reuse step reaches $11.1\times$ its
isolated-reuse counterpart. The corresponding amplification is
$7.0\times$ for value reuse and $7.5\times$ for joint key--value reuse.
Reusing only keys or attention probabilities gives factors of
$1.6\times$ and $2.0\times$, respectively. Repeated approximations therefore exhibit different interactions
across the cached quantities.

The distinction predicts different placement preferences. On 30 held-out
scenes with a budget of five reuse steps, spreading output reuse across
the trajectory reduces rotation deviation from $0.710^\circ$ for the
late concentrated layout to $0.374^\circ$, winning on 25 scenes. The
spread-out placement also wins on 27 scenes for value reuse and 22 for
key--value reuse. For attention-probability reuse, the concentrated
layout wins on 22 scenes. The preferred reuse layout thus varies
with the cached quantity in these comparisons. Across the tested cache quantities, the reported Spearman correlations between stability and single-reuse local error are 0.95--1.00. The spacing comparison evaluates the accumulated effect directly on the final poses.

\subsection{Precision and perturbation magnitude}
\label{app:precision}

We repeat the propagation experiments in FP32 using the same scenes,
intervention plans, and sampled perturbations as the bfloat16 runs. Conditions are grouped by their median FP32 relative injection
magnitude. Within each group, we summarize the ratios of FP32 to
bfloat16 condition-level retained fractions. The median ratio is
0.35 below 1\% injection magnitude and 0.98 at magnitudes of at least
5\%. Conditions with nonpositive bfloat16 values are excluded.
Repeating a deterministic clean forward pass gives zero state deviation,
while changed inputs can induce different rounding along the perturbed
trajectory. FP32 comparisons resolve the resulting inflation of
small-perturbation gains.

Across scales 0.25, 0.5, and 1, the maximum-to-minimum retained-fraction
ratio is below 1.10 in 54 of the 60 evaluated conditions.
Larger perturbations exhibit finite-amplitude effects: for
KV pooling at step 12 in \dejaview, the retained fractions at
scales 0.25, 0.5, 1, and 2 are 0.77, 0.79, 0.85, and 1.25.
For middle-depth KV pooling in VGGT they are 4.3, 4.8, 6.3,
and 9.8. The perturbation scale is therefore part of the intervention
specification.

\subsection{Recovery of poses and task accuracy}
\label{app:recovery}

Across the recorded intervention trajectories, the median gain is 0.926 for the original \dejaview model and
0.924 for the adapted 14-step model. In the latter, KV-pooling
interventions at steps 2 and 7 retain fractions of 0.47 and 0.74;
reuse interventions retain 0.36 and 0.60.
For task recovery, we measure the final AUC@$30^\circ$ loss relative
to the loss at the first logged post-intervention readout, retaining
scenes whose initial loss exceeds $10^{-4}$. Trajectory recovery uses
the corresponding ratio of output rotation deviations, with initial
values at or below $10^{-9}$ excluded. A trajectory recovers without
task recovery when its rotation deviation decreases while its final
task loss remains at least as large as its initial positive loss.
Ground-truth curves summarize each trajectory by scene medians;
paired differences are computed within scenes before taking their
median. In the logged 14-step protocol, intervention at step 12 has
its first post-intervention readout at the final output and supplies
no additional recovery interval.

Rotation-deviation recovery and task recovery remain distinct
(\cref{tab:task_recovery}). After coarse-grid processing at step 12,
15 of 65 scenes reduce their rotation deviation while retaining at
least their initial positive AUC@$30^\circ$ loss. The corresponding
count for KV pooling is 12. These cases connect state-propagation
analysis to the need for direct task-level evaluation.

\subsection{Linearized sensitivity of the prediction heads}
\label{app:decoder}

At matched final-state norms, decoder JVPs give real-to-random response
ratios of 1.36 for rotation, 1.85 for camera centers, and 1.24 for
depth. These local sensitivity differences agree in direction with
the finite decoding experiments in \cref{sec:propagation}. The
reported scene-wise correlation of the combined linear prediction
with the observed rotation comparison is 0.13. We use the decoder
analysis to explain directional sensitivity and evaluate schedule
quality directly with pose and geometry metrics.

\subsection{Predicting the cost of unseen operators and layouts}
\label{app:prediction}

The prediction target is the final rotation deviation from the clean reconstruction. We evaluate operator rankings at individual steps and layout selection under a fixed budget separately.

The predictor is multiplicative. For an approximation applied at step
$u$, let $\epsilon_u$ be its local error, the relative error of the
global-attention output at that step; for a coarse-grid step it is the
state deviation right after the step. The cost $c$ is the
rotation deviation of the final prediction from the clean prediction.
The predicted cost is $\epsilon_u\exp(h(u))$, where the position
profile $h(u)$ is the median of $\log(c/\epsilon_u)$ over the
discovery scenes and over five perturbations applied at step $u$: KV
pooling, the coarse grid, output reuse, AVGGT-style key subsampling,
and random noise. The profile rises from $-2.1$ at the first step to
$-0.8$ at the last. The three reduced predictors keep only
$\exp(h(u))$, only $\epsilon_u$, or only the change of the
global-attention output between consecutive steps. For each held-out
approximation type we take the median predicted and measured cost over
the 30 test scenes at each of the 16 steps, and report the Spearman
correlation across steps. For a budget of several approximate steps,
the prediction sums the per-step predictions, using the median
discovery $\epsilon_u$, over 12 candidate layouts; regret is the
measured cost of the predicted best layout relative to the measured
best.

We fit a cost predictor on the 35 even-indexed scenes and evaluate it
on the 30 odd-indexed scenes. Across three held-out approximation
types, the predictor achieves rank correlations of 0.84--0.89 with
measured cost (\cref{tab:operator_prediction}). The useful predictors
vary with the operation: position alone matches the full predictor
for pooling on half the heads, while local error alone gives the
highest correlation for attention-probability reuse.

For coarse-grid and KV-pooling budgets of three and six approximate
steps, the predictor selects the early-prefix layout with zero regret
within the evaluated candidate sets. Position alone selects the same
layouts. Cross-step reuse exhibits a different pattern: with three,
four, and five reuse steps, the full predictor has rank correlations
of 0.80, 0.33, and 0.09, and selection regrets of 12\%, 73\%, and
95\%, respectively. Prediction degrades as reuse interventions are
composed, motivating direct evaluation of complete reuse layouts.

\begin{table}[!htbp]
  \centering
  \small
  \setlength{\tabcolsep}{3pt}
  \caption{Rank correlation between predicted and measured cost on
  held-out approximation types. Pos.: position only; Local: local
  error only; Stab.: stability only.}
  \label{tab:operator_prediction}
  \begin{tabular}{lrrrr}
    \toprule
    Approximation & Full & Pos. & Local & Stab. \\
    \midrule
    Top-256 sparse & 0.89 & 0.86 & 0.62 & $-1.00$ \\
    Probability reuse & 0.89 & $-0.80$ & 0.92 & 0.92 \\
    Half-head pooling & 0.84 & 0.84 & $-0.19$ & $-0.98$ \\
    \bottomrule
  \end{tabular}
\end{table}

\subsection{Placement across budgets and view counts}
\label{app:transfer}

\Cref{tab:placement_transfer} varies the loop budget and number of views. The paired ScanNet++ experiment in \cref{tab:long_sequence_placement} changes the number of views of the same scene, allowing the effect of view count to be examined within each scene.

Early placement gives the smallest AUC loss for four pooling steps
in the original 12- and 16-step models, including the 100- and
300-view evaluations. At eight steps, uniform placement performs best,
with a 14.5-point loss compared with 17.7 points for early placement.
For the adapted eight-step configuration, late placement gives the
smallest AUC loss (8.64 points), followed by early placement (10.51)
and uniform placement (17.54). At 12 and 16 steps, early placement remains best
under the standard view setting and at 16 steps with 100 views.
A follow-up evaluation on 16 ScanNet++ scenes compares the layouts
using 151--300 views and paired 100-view inputs
(\cref{tab:long_sequence_placement}). Single-intervention position
correlations with output rotation deviation remain positive across
the earlier settings, while multi-intervention task losses also depend
on the scene and the computation already present in the adapted schedule.

For five reuse steps in the original 16-step model, a dispersed layout
reduces the AUC loss from 10.4 to 3.9 points. The same comparison
changes the loss from 8.8 to 3.3 points at 100 views. The tested
layouts coincide at 8 and
12 steps, where the reuse budgets are three and five, respectively.
These budgets provide no distinct spacing comparison.

\noindent\textbf{Paired long-sequence evaluation.}
We evaluate 16 ScanNet++ scenes with 151--300 input views, including
11 scenes with exactly 300 views. With four KV-pooling steps, the
original model loses 1.2 AUC@$3^\circ$ points under early placement,
compared with 7.2 for uniform and 9.9 for late placement
(\cref{tab:long_sequence_placement}). This extends the observed early
placement advantage to the evaluated long sequences.

The adapted model exhibits heterogeneous scene-level responses.
Early placement loses 16.0 points on average, compared with 11.6
for uniform and 9.4 for late placement. We define a failure as an AUC@$3^\circ$ drop greater than 20 points.
Six scenes lose 30--46 points under early placement, while eight
remain nearly unchanged. Early
placement still gives the smallest rotation deviation, so the ranking
depends on the evaluation metric. Paired 100-view and long-sequence
evaluations on these same scenes give similar outcomes, with the
same scenes exhibiting large losses. In the earlier 32-view study,
failures occur in five of the 13 ScanNet++ scenes and in none of the
other four datasets. The observed failures are thus concentrated in
a subset of scenes and persist across the paired view-count settings.

\begin{table*}[!htbp]
  \centering
  \small
  \setlength{\tabcolsep}{4pt}
  \caption{Placement across budgets, view counts, and adaptation.
  Entries under pooling and reuse are AUC@$3^\circ$ changes in
  percentage points; larger values indicate smaller losses.
  $\rho$ correlates single-pooling output rotation deviation with position.
  Each row uses its own unmodified configuration as the reference.
  Reuse budgets are five except at $K=8$, where they are three.
  The consecutive and dispersed candidates coincide at $K=8$ and $K=12$.
  Dashes denote unavailable comparisons.}
  \label{tab:placement_transfer}
  \begin{tabular}{lrrrrrrr}
    \toprule
    Setting & Scenes & $\rho$ & \multicolumn{3}{c}{Four pooling steps}
    & \multicolumn{2}{c}{Reuse} \\
    \cmidrule(lr){4-6}\cmidrule(lr){7-8}
    & & & Early & Uniform & Late & Consecutive & Dispersed \\
    \midrule
    Original, $K=8$ & 65 & 0.86 & $-17.7$ & $-14.5$ & $-16.0$ & $-39.2$ & $-39.2$ \\
    Original, $K=12$ & 65 & 0.97 & $-5.0$ & $-8.2$ & $-13.4$ & $-50.1$ & $-50.1$ \\
    Original, $K=16$ & 65 & 0.97 & $-1.8$ & $-5.5$ & $-9.5$ & $-10.4$ & $-3.9$ \\
    Original, $K=16$, 100 views & 20 & 0.96 & $-1.8$ & $-5.7$ & $-7.6$ & $-8.8$ & $-3.3$ \\
    \midrule
    Adapted, $K=8$ & 65 & 0.64 & $-10.5$ & $-17.5$ & $-8.6$ & $-42.2$ & $-42.2$ \\
    Adapted, $K=12$ & 65 & 0.83 & $-4.0$ & $-9.2$ & $-7.8$ & \NA & \NA \\
    Adapted, $K=16$ & 65 & 0.92 & $-3.8$ & $-7.7$ & $-7.3$ & \NA & \NA \\
    Adapted, $K=16$, 100 views & 20 & 0.91 & $-5.5$ & $-8.8$ & $-6.3$ & \NA & \NA \\
    \bottomrule
  \end{tabular}
\end{table*}

\begin{table}[!htbp]
  \centering
  \small
  \setlength{\tabcolsep}{5pt}
  \caption{Four KV-pooling interventions on 16 ScanNet++ scenes with
  151--300 views, including 11 scenes at 300 views. Entries are mean
  AUC@$3^\circ$ changes in percentage points relative to each model's
  unmodified configuration.}
  \label{tab:long_sequence_placement}
  \begin{tabular}{lrrr}
    \toprule
    Model & Early & Uniform & Late \\
    \midrule
    Original & $-1.2$ & $-7.2$ & $-9.9$ \\
    Adapted & $-16.0$ & $-11.6$ & $-9.4$ \\
    \bottomrule
  \end{tabular}
\end{table}

\subsection{Interaction between successive interventions}
\label{app:interactions}

We combine an initial pooling or noise intervention with windows of
one, two, or four modified updates. For final deviation $E$, we
summarize the interaction between an initial intervention $A$ and a
window $B$ as
\begin{equation}
  I(A,B)=E(A+B)-E(A)-E(B),
  \label{eq:interaction}
\end{equation}
where $A+B$ denotes their joint application. Most reported interactions
for freezing values are nonpositive. The strongest positive interaction
among the tested value-cache conditions occurs when values are cached
for four steps after pooling at step 8: $I=0.127$, with a 95\%
confidence interval of $[0.110,0.161]$. The corresponding four-step
value-freezing condition gives $-0.033$ $[-0.062,-0.012]$.
Interactions depend on the intervention, its position, and the way
cached quantities are maintained. We use these operator-specific
responses to characterize reuse composition.

\section{Training and evaluation of \ours}
\label{app:experiments}

We compare \ours with the released model and a full-block baseline adapted by distillation. The budget sweep measures the effect of changing compute within a checkpoint. Placement controls hold the training setup fixed and change the positions of the approximate updates.

\subsection{Adaptation and evaluation protocol}
\label{app:setup}

\noindent\textbf{Datasets and metrics.}
We follow the released \dejaview evaluation
protocol~\cite{burzio2026dejaview} on five benchmarks: 22 DTU scans,
13 ETH3D scenes, 18 test sequences from 7-Scenes, 50 ScanNet++ scenes
from the \texttt{nvs\_sem\_val} split, and 50 nuScenes evaluation
scenes. DTU uses all 49 views; ETH3D uses all available views, ranging
from 14 to 76 in our evaluation. We sample every eleventh frame for
7-Scenes and use up to 100 and 50 frames for ScanNet++ and nuScenes,
respectively. Long-sequence evaluation uses every third frame for
7-Scenes, approximately 334 views per 1,000-frame sequence, and 300
frames for ScanNet++.

Camera pose accuracy is measured by AUC@$3^\circ$ and AUC@$30^\circ$,
using the maximum of the relative rotation and translation angular
errors for each camera pair. For pointmaps, we apply Sim(3) alignment
to ground truth and report relative $\ell_2$ error and the inlier ratio
at a 3\% relative-error threshold. Evaluation uses the official
alignment and validity filtering.

\noindent\textbf{Input and implementation.}
The input long edge is 504 pixels. The resulting image sizes are
$378\times504$ for DTU and 7-Scenes, $336\times504$ for ETH3D and
ScanNet++, and $280\times504$ for nuScenes. Inference processes one
scene at a time using bfloat16 autocasting and PyTorch scaled
dot-product attention, with the FlashAttention-2 backend on A100.
Pose evaluation uses float64 arithmetic.

\noindent\textbf{Baselines and training fairness.}
The reference is the official 117M-parameter \dejaview checkpoint
evaluated with $K=16$. We also evaluate the same checkpoint at
$K\in\{8,10,12,14\}$ to measure the trade-off from reducing iteration
count. Our budget sweep uses a single adapted checkpoint. Mechanism
experiments in \cref{sec:analysis} use the released weights,
and the accelerated model uses the adaptation recipe in
\cref{sec:adaptation}. The main-table speedup for each configuration
is its $K=16$ reference latency divided by its measured latency under
the corresponding evaluation setting.

\noindent\textbf{Adaptation recipe.}
All adapted models, including the plain baseline and the ablation
checkpoints, share one recipe. The teacher is the released model at
$K=16$ and no labels are used. Each training step draws one scene of 4
to 24 views and one budget $K$, runs the student with its approximate
schedule, and decodes the final state with the frozen heads. With
teacher quantities marked by a star, the loss is
\begin{equation}
\begin{aligned}
  \mathcal L ={}& \overline{|\log d-\log d^\star|}
    +\frac{\overline{|X-X^\star|}}{s^\star}
    +\frac{\overline{|r-r^\star|}}{\overline{|r^\star|}}
    +\overline{|p-p^\star|}\\
  &+10\,\theta(r,r^\star)
    +0.2\,(\rho_{\mathrm{rot}}+\rho_{\mathrm{trans}})
    +0.5\,\mathcal L_{\mathrm{fit}},
\end{aligned}
\label{eq:distill_loss}
\end{equation}
where bars denote means over pixels and views, $d$ is depth, $X$ the
point map and $s^\star$ the mean norm of the teacher's points, $r$ the
ray map, and $p$ the pose encoding. $\theta$ is the mean angle in
radians between student and teacher ray directions.
$\rho_{\mathrm{rot}}$ and $\rho_{\mathrm{trans}}$ are the mean
disagreements, in degrees, of the relative rotation and of the
relative translation direction over all view pairs.
$\mathcal L_{\mathrm{fit}}$ applies to KV-pooling steps: it is the
squared error between the pooled and the exact attention output of the
same queries, normalized by the squared norm of the exact output and
summed over the pooling steps. We train for 12,000 steps with AdamW,
no weight decay, gradient clipping at norm 1, 300 warmup steps, and
cosine decay. The learning rate is $5\times10^{-4}$ for the LoRA
anchors, $2\times10^{-3}$ for the multistep weights, and $10^{-3}$ for
the KV-pooling parameters.

\noindent\textbf{Distillation data and sampling.}
The distillation images are the subset described in
\cref{sec:adaptation}, about 3\% of the images of the training splits
of the \dejaview training datasets. Each optimizer step uses a single
scene, drawn from a dataset chosen with equal probability, and its
number of views is drawn from 4 to 24. The budget $K$ of the step is
drawn uniformly from the set of budgets of the checkpoint. The tier
checkpoint instead draws one of the three schedules of
\cref{sec:schedule} uniformly at every step, so one set of anchors
serves the quality, balanced, and fast settings. One scene of each
dataset is held out from training and used to monitor the distillation
loss. The student runs its schedule during training, including the
reuse steps, and gradients pass through all steps.

\noindent\textbf{Multistep update.}
Let $D_t=f_\theta(z_t;\tau_t,\tau_{t+1})-z_t$ be the update of the
block at step $t$, computed with the operator of that step. The state
advances as
\begin{equation}
  z_{t+1}=z_t+D_t+c_1(\tau_t)\odot D_{t-1}+c_2(\tau_t)\odot D_{t-2},
  \label{eq:multistep}
\end{equation}
where $c_1,c_2\in\mathbb R^{D}$ are per-channel weights and $\odot$ is
the channel-wise product. Each weight is stored at seven anchors and
linearly interpolated at $\tau_t$. The weights are initialized to
zero, so training starts from the plain loop. Terms whose update does
not exist are dropped: the first step uses no history, and the
history is cleared at each change between the coarse and the full
grid. Successive coarse steps retain their update history on the coarse grid. The channel gates act inside $f_\theta$ and are unchanged.

\noindent\textbf{Learned KV pooling.}
In a KV-pooling step, each pooled key and value is a weighted mean of
the tokens of its $2\times2$ window. Head $h$ holds a score vector
$u_h$ and a scalar $b_h$. The weights of a window are the softmax of
the scores $\langle k_i,u_h\rangle$ of its keys $k_i$, and the same
weights pool the values. The bias added to the attention logit of a
pooled key is $\log n+b_h$, where $n$ is the number of tokens in its
window. Both $u_h$ and $b_h$ start at zero, which gives mean pooling
with bias $\log n$, so training starts from the unlearned operator.
The camera and register tokens stay unpooled and receive no bias.

\noindent\textbf{Attention-output cache.}
The cached tensor $A_t$ of \cref{eq:attention_reuse} is the output of
the output projection of global attention, taken before LayerScale and
before the time-conditioned gate. A reuse step applies its own
LayerScale and its own gate to the cached tensor, so the time
conditioning of the step is kept. A reuse step needs a full-resolution
attention output from the step before it. Every schedule therefore
places a reuse step directly after a full or a KV-pooled step, and
never after a coarse-grid step or another reuse step.

\noindent\textbf{Parameter accounting.}
The adapted model stores approximately 120.6M parameters, comprising
117.1M base parameters, 3,538,944 parameters in nine LoRA anchor sets,
10,752 multistep weights (two weights per channel at seven anchors,
for 768 channels), and the per-head scores and biases of learned KV
pooling.
Parameter counts refer to this compact representation. At inference,
precomputed dense projections materialize the step-specific updates, and
their additional storage is accounted for in the runtime memory footprint.

\noindent\textbf{Timing protocol.}
We measure latency on an exclusively allocated A100-SXM4-80GB GPU.
Each measurement uses two warmup runs followed by five timed runs,
with CUDA synchronization before and after each forward pass; we
report the median. Timing covers the encoder, recurrent updates,
prediction heads, cache accesses, and schedule transitions. Image
loading, preprocessing, and evaluation alignment are outside the timed
region. Step-specific merged projections are precomputed before timing;
their resident storage contributes to inference memory usage. Compilation
with \texttt{torch.compile} is disabled. \dejaview and the plain
baseline are timed with the original implementation of the block, and
our schedules with our optimized implementation.

The results in \cref{tab:main} use the dataset-specific view counts
listed in the table for both accuracy and timing.

\subsection{Operating points and comparisons at matched latency}
\label{app:main_results}

\Cref{tab:budget_curve} lists the measured operating points. For \cref{tab:matched_latency}, we interpolate each baseline between measured latencies and compare its accuracy with the accelerated model at the same latency. The plain baseline and the sweep use the following training and scheduling choices.

\noindent\textbf{Plain baseline.}
The plain baseline is adapted with the recipe of our models
(\cref{app:setup}): the same nine rank-16 LoRA anchor sets on the same
eight layers, the same learned multistep update, and the same teacher,
loss, data, optimizer, and 12,000 training steps. All of its steps are
exact, and training samples $K\in\{4,5,6,8,10,12,14,16\}$, so one
checkpoint gives the whole plain curve. It measures what the same
adaptation gives to a shorter loop without our operators.

\noindent\textbf{Schedules of the sweep.}
The sweep checkpoint runs $\mathrm{C}^{K-8}\mathrm{P}^{4}\mathrm{FRFR}$
at every budget, which gives $\mathrm{P}^{4}\mathrm{FRFR}$ at $K=8$
and $\mathrm{C}^{8}\mathrm{P}^{4}\mathrm{FRFR}$ at $K=16$. Training
samples $K\in\{8,10,12,14,16\}$ with this schedule. At $K=14$ the
schedule equals that of the fast tier. The two rows are different
checkpoints: the sweep checkpoint is trained on this schedule family
alone, and the tier checkpoint is shared by the quality, balanced, and
fast schedules. They reach 52.00 and 51.02 AUC@$3^\circ$ at 0.790 and
0.787 seconds.

Reducing the original model from 16 to 14 steps gives
$1.10$--$1.12\times$ speedup. At the same 14-step count, the balanced
tier is $1.31$--$1.53\times$ faster than the original model, and the
fast tier is $1.66$--$2.24\times$ faster. These comparisons separate
the reduction in update cost from the reduction in iteration count.
Budget-curve summaries weight the five datasets equally for both
accuracy and latency. Matched-latency comparisons linearly interpolate
the reported averages within each baseline's measured latency range. Speedups retain the official
16-step model as the numerator. Configurations timed in separate runs
are rescaled by the 16-step reference timed in the same run; the
correction is at most 1.1\%.

The balanced tier obtains AUC@$3^\circ$ of 55.3 at 1.106 seconds,
compared with 54.1 at 1.261 seconds for plain $K=10$. The quality
tier obtains 56.3 at 1.380 seconds, compared with 55.8 at 1.449 seconds
for plain $K=12$. Geometry also favors the quality tier in this
comparison: its inlier score is 78.9 versus 78.5, and AbsRel is
0.0317 versus 0.0324. These measured points complement the
interpolated matched-latency comparisons.

\noindent\textbf{Comparison at matched latency.}
\Cref{tab:matched_latency} compares each configuration with linear
interpolations of the original and plain baseline curves at the same
latency, within their measured ranges. Across the five single-checkpoint
budgets, all four macro-average metrics improve over both interpolated
baselines. At $K=14$, AUC@$3^\circ$ improves by 33.8 points over the
original curve and by 20.9 over the plain curve; AbsRel decreases by
0.0265 and 0.0188, respectively.

The balanced tier improves AUC@$3^\circ$ by 9.1 points over the
original curve and by 4.7 over the plain curve, with improvements in
AUC@$30^\circ$, inlier score, and AbsRel as well. The quality tier
has smaller gains: AUC@$3^\circ$ improves by 2.0 and 1.1 points,
its inlier score by 0.6 and 1.0 points, and AbsRel decreases by
$0.4\times10^{-3}$ and $1.3\times10^{-3}$. These
comparisons characterize the quality available at each operating
latency; speedup factors in the table use the fixed original $K=16$
reference.

\begin{table*}[!htbp]
  \centering
  \small
  \setlength{\tabcolsep}{4pt}
  \caption{Matched-latency comparisons using equal-weight dataset
  macro-averages. Each pair is ours minus the original /
  plain baseline linearly interpolated within its measured latency range.
  AUC and inlier changes are percentage points; AbsRel changes are
  multiplied by 1,000. Speedup uses the budget experiment's
  original $K=16$ latency (1.804 seconds) as numerator.}
  \label{tab:matched_latency}
  \begin{tabular}{lrrrrrr}
    \toprule
    Configuration & Time (s) & Speedup & $\Delta$AUC@$3^\circ$ & $\Delta$AUC@$30^\circ$ & $\Delta$Inlier & $\Delta$AbsRel $\times10^3$ \\
    \midrule
    \ours, sweep ($K=8$) & 0.719 & $2.51\times$ & $+31.5 / +19.8$ & $+15.4 / +9.0$ & $+21.2 / +14.4$ & $-40.5 / -23.1$ \\
    \ours, sweep ($K=10$) & 0.744 & $2.42\times$ & $+35.4 / +23.2$ & $+15.4 / +9.8$ & $+22.4 / +16.5$ & $-37.8 / -24.1$ \\
    \ours, sweep ($K=12$) & 0.767 & $2.35\times$ & $+35.2 / +22.6$ & $+14.1 / +9.1$ & $+20.8 / +15.8$ & $-32.7 / -22.3$ \\
    \ours, sweep ($K=14$) & 0.790 & $2.28\times$ & $+33.8 / +20.9$ & $+12.4 / +7.9$ & $+18.7 / +14.2$ & $-26.5 / -18.8$ \\
    \ours, sweep ($K=16$) & 0.813 & $2.22\times$ & $+30.9 / +18.4$ & $+11.1 / +7.1$ & $+16.8 / +12.8$ & $-23.1 / -16.3$ \\
    \midrule
    \ours, quality & 1.380 & $1.31\times$ & $+2.0 / +1.1$ & $+0.4 / +0.6$ & $+0.6 / +1.0$ & $-0.4 / -1.3$ \\
    \ours, balanced & 1.106 & $1.63\times$ & $+9.1 / +4.7$ & $+2.2 / +1.8$ & $+3.8 / +3.5$ & $-3.9 / -4.2$ \\
    \ours, fast & 0.787 & $2.29\times$ & $+33.2 / +20.3$ & $+12.5 / +8.1$ & $+18.8 / +14.4$ & $-26.8 / -19.2$ \\
    \bottomrule
  \end{tabular}
\end{table*}

\begin{table*}[!htbp]
  \centering
  \small
  \setlength{\tabcolsep}{4pt}
  \caption{Complete budget evaluation with equal dataset weights.
  Entries report averaged results. AUC and inlier scores are
  percentages; AbsRel is multiplied
  by 1,000. Single-checkpoint curves and tier operating points are separate groups.}
  \label{tab:budget_curve}
  \begin{tabular}{lrrrrr}
    \toprule
    Configuration & Time (s) & AUC@$3^\circ$ & AUC@$30^\circ$ & Inlier & AbsRel $\times10^3$ \\
    \midrule
    Original ($K=4$) & 0.692 & 8.05 & 68.97 & 42.65 & 92.63 \\
    Original ($K=5$) & 0.786 & 17.72 & 78.03 & 57.06 & 61.55 \\
    Original ($K=6$) & 0.879 & 28.57 & 83.54 & 65.15 & 47.86 \\
    Original ($K=8$) & 1.065 & 44.66 & 88.78 & 73.62 & 37.06 \\
    Original ($K=10$) & 1.251 & 51.95 & 90.71 & 77.04 & 33.21 \\
    Original ($K=12$) & 1.436 & 55.30 & 91.40 & 78.89 & 31.65 \\
    Original ($K=14$) & 1.622 & 56.67 & 91.72 & 79.56 & 31.12 \\
    Original ($K=16$) & 1.804 & 57.20 & 91.85 & 79.95 & 30.96 \\
    \midrule
    Plain ($K=4$) & 0.698 & 20.02 & 76.59 & 51.18 & 70.25 \\
    Plain ($K=5$) & 0.791 & 31.25 & 82.75 & 61.97 & 53.02 \\
    Plain ($K=6$) & 0.885 & 40.15 & 86.38 & 68.34 & 43.70 \\
    Plain ($K=8$) & 1.074 & 49.88 & 89.34 & 74.27 & 37.12 \\
    Plain ($K=10$) & 1.261 & 54.08 & 90.61 & 76.91 & 33.83 \\
    Plain ($K=12$) & 1.449 & 55.83 & 91.32 & 78.48 & 32.44 \\
    Plain ($K=14$) & 1.632 & 56.71 & 91.65 & 79.10 & 31.69 \\
    Plain ($K=16$) & 1.827 & 57.11 & 91.76 & 79.46 & 31.62 \\
    \midrule
    \ours, sweep ($K=8$) & 0.719 & 42.35 & 86.92 & 67.97 & 43.30 \\
    \ours, sweep ($K=10$) & 0.744 & 48.74 & 89.39 & 73.02 & 37.67 \\
    \ours, sweep ($K=12$) & 0.767 & 50.93 & 90.26 & 74.94 & 35.19 \\
    \ours, sweep ($K=14$) & 0.790 & 52.00 & 90.61 & 76.04 & 34.46 \\
    \ours, sweep ($K=16$) & 0.813 & 51.73 & 90.67 & 76.20 & 34.57 \\
    \midrule
    \ours, quality ($K=16$) & 1.380 & 56.30 & 91.63 & 78.92 & 31.70 \\
    \ours, balanced ($K=14$) & 1.106 & 55.34 & 91.39 & 78.19 & 32.34 \\
    \ours, fast ($K=14$) & 0.787 & 51.02 & 90.54 & 75.89 & 34.63 \\
    \bottomrule
  \end{tabular}
\end{table*}

\subsection{Placement with matched adaptation}
\label{app:schedule_ablation}

The comparison in \cref{tab:schedules} uses paired differences on the same scenes. Each row compares the proposed layout with a control at the same loop count.

We compare schedules that differ only in the positions of approximate
operations, using the same training setup. Every control keeps the
operator counts of our schedule $\mathrm{C}^{K-8}\mathrm{P}^{4}\mathrm{FRFR}$
at the same $K$: $K-8$ coarse-grid steps, four KV-pooling steps, two
full steps, and two reuse steps. Each control has its own adapter,
trained with the recipe of \cref{app:setup}.

\noindent\textbf{Controls.}
The \emph{uniform} control spreads the steps of each operator evenly
over the trajectory. The \emph{reversed} control runs our schedule
backwards, $\mathrm{FRFR}\,\mathrm{P}^{4}\mathrm{C}^{K-8}$, so reuse
comes first and the coarse grid last. The \emph{stability} control
follows the rule of diffusion caching, which approximates where
features change least. On clean runs of the released model on
training scenes, we measure the relative change of the
global-attention output between consecutive steps and rank the steps
from most to least stable. The most stable steps receive reuse, the
next KV pooling, the next the coarse grid, and the least stable steps
stay full. The \emph{random} control permutes the steps of our
schedule at random. We draw eight such layouts, train an adapter for
2,000 steps on each, and keep the layout with the lowest distillation
loss on held-out training scenes, averaged over the budgets; this
layout is then trained with the full recipe. No benchmark image enters
the stability ranking or the selection. In all controls a reuse step
directly follows a full or a KV-pooled step, as required by the cache
(\cref{app:setup}).

\noindent\textbf{Results.}
Comparisons are paired by scene
across the 153-scene evaluation set. The denominator in the wins column of \cref{tab:schedules} gives the number of paired scenes for that row.

\Cref{tab:schedules} reports differences relative to each control.
At 14 steps, our schedule improves AUC@$3^\circ$ by 2.62 points over
uniform placement, 4.90 over reversed placement, 1.53 over the
stability-based rule, and 1.96 over the selected random schedule. Across 10, 12, 14, and 16 steps, all 16 AUC@$3^\circ$
comparisons have 95\% confidence intervals above zero. Geometry
improvements also appear across the comparisons, with intervals
crossing zero for some stability and random controls. The placement
advantage therefore persists after matched adaptation.

Measured latency differs by at most 0.012 seconds between our schedule
and the uniform, reversed, and random controls at the same loop count.

\begin{table*}[!htbp]
  \centering
  \footnotesize
  \setlength{\tabcolsep}{3pt}
  \caption{Effect of approximation placement under matched adaptation.
  Entries report ours minus control with 95\% confidence intervals.
  AUC and inlier differences are percentage points; AbsRel differences
  are multiplied by 1,000. Negative AbsRel differences favor our schedule.
  Wins count scenes with higher AUC@$3^\circ$. Random (val.) is the best of eight random layouts by distillation loss on held-out training scenes.}
  \label{tab:schedules}
  \begin{tabular}{rlrrrrr}
    \toprule
    $K$ & Control & $\Delta$AUC@$3^\circ$ & $\Delta$AUC@$30^\circ$
    & $\Delta$Inlier & $\Delta$AbsRel $\times 10^3$ & Wins \\
    \midrule
    10 & Uniform & $+2.74\,[+1.93,+3.66]$ & $+1.11\,[+0.60,+1.71]$ & $+1.88\,[+0.44,+3.62]$ & $-3.03\,[-6.24,-0.54]$ & 110/153 \\
    10 & Reversed & $+4.13\,[+3.35,+4.92]$ & $+1.67\,[+1.06,+2.51]$ & $+2.23\,[+1.14,+3.74]$ & $-3.32\,[-6.22,-1.21]$ & 111/152 \\
    10 & Stability & $+0.89\,[+0.38,+1.40]$ & $+0.46\,[+0.14,+0.85]$ & $+0.75\,[-0.37,+2.21]$ & $-0.77\,[-2.63,+0.52]$ & 106/153 \\
    10 & Random (val.) & $+3.55\,[+2.96,+4.15]$ & $+1.12\,[+0.71,+1.61]$ & $+2.34\,[+1.45,+3.37]$ & $-2.87\,[-4.42,-1.53]$ & 126/153 \\
    \midrule
    12 & Uniform & $+3.28\,[+2.76,+3.79]$ & $+1.27\,[+0.61,+2.29]$ & $+2.47\,[+1.23,+4.00]$ & $-3.15\,[-5.48,-1.17]$ & 119/153 \\
    12 & Reversed & $+4.01\,[+3.30,+4.73]$ & $+1.62\,[+0.80,+2.99]$ & $+3.07\,[+1.68,+5.01]$ & $-4.35\,[-7.94,-1.45]$ & 116/153 \\
    12 & Stability & $+1.06\,[+0.61,+1.52]$ & $+0.48\,[+0.21,+0.76]$ & $+1.15\,[+0.13,+2.51]$ & $-1.06\,[-2.56,-0.03]$ & 102/153 \\
    12 & Random (val.) & $+1.79\,[+1.18,+2.34]$ & $+0.78\,[+0.24,+1.48]$ & $+1.93\,[+0.73,+3.58]$ & $-2.68\,[-5.03,-0.71]$ & 112/153 \\
    \midrule
    14 & Uniform & $+2.62\,[+1.89,+3.35]$ & $+0.95\,[+0.27,+1.90]$ & $+2.54\,[+1.04,+4.64]$ & $-3.31\,[-6.55,-0.70]$ & 113/153 \\
    14 & Reversed & $+4.90\,[+4.11,+5.71]$ & $+1.70\,[+0.77,+3.13]$ & $+4.14\,[+2.51,+6.16]$ & $-4.70\,[-7.97,-1.99]$ & 124/151 \\
    14 & Stability & $+1.53\,[+1.00,+2.09]$ & $+0.39\,[+0.09,+0.72]$ & $+1.29\,[+0.55,+2.09]$ & $-0.88\,[-1.79,-0.10]$ & 99/153 \\
    14 & Random (val.) & $+1.96\,[+1.29,+2.57]$ & $+0.64\,[+0.13,+1.27]$ & $+1.80\,[+0.39,+3.57]$ & $-1.48\,[-3.60,+0.15]$ & 110/153 \\
    \midrule
    16 & Uniform & $+2.04\,[+1.28,+2.81]$ & $+0.77\,[+0.14,+1.62]$ & $+2.13\,[+0.82,+3.71]$ & $-2.29\,[-4.36,-0.48]$ & 103/152 \\
    16 & Reversed & $+5.61\,[+4.72,+6.58]$ & $+1.64\,[+0.79,+2.85]$ & $+4.97\,[+3.36,+7.03]$ & $-4.82\,[-7.63,-2.46]$ & 123/153 \\
    16 & Stability & $+1.30\,[+0.75,+1.85]$ & $+0.26\,[-0.02,+0.52]$ & $+1.16\,[+0.58,+1.80]$ & $-0.68\,[-1.45,+0.17]$ & 102/153 \\
    16 & Random (val.) & $+5.85\,[+5.04,+6.75]$ & $+1.54\,[+0.95,+2.38]$ & $+4.06\,[+2.65,+5.70]$ & $-3.48\,[-5.20,-1.93]$ & 124/153 \\
    \bottomrule
  \end{tabular}
\end{table*}

\subsection{Which operations contribute to the speedup?}
\label{app:components}

\Cref{tab:components,tab:component_tradeoffs} enumerate the eight configurations. Disabling an operator restores the exact computation at its assigned steps.

Each of the eight combinations has its own adapter, trained with the
recipe of \cref{app:setup} and $K\in\{8,10,12,14,16\}$. All
combinations keep the positions of the full schedule
$\mathrm{C}^{K-8}\mathrm{P}^{4}\mathrm{FRFR}$ and run the disabled
operators as exact steps. At $K=14$ the coarse-grid steps are steps 1
to 6, the KV-pooling steps are steps 7 to 10, and the reuse steps are
steps 12 and 14. The row with all operators disabled is the plain
baseline at $K=14$, and the full combination is the sweep checkpoint
at $K=14$.

Relative to the configuration with all operations disabled, reuse
alone reduces macro-average AUC@$3^\circ$ by 0.19 points and average
latency from 1.629 to 1.497 seconds. Coarse-grid processing reduces
AUC by 2.47 points and latency to 1.173 seconds. The full combination
reaches 0.790 seconds, giving $2.28\times$ speedup over the official
16-step reference measured in this experiment, with AUC 52.00
compared with 57.20. The operations occupy different points on the
accuracy--latency trade-off.

Geometry reveals additional interactions. On ETH3D, combining pooling
and reuse gives AbsRel 0.0449, compared with 0.0305 for pooling alone
and 0.0291 for reuse alone. The full combination gives 0.0370 on this
dataset, alongside lower AUC@$3^\circ$. We therefore evaluate the
component combinations jointly in pose, geometry, and latency.

\begin{table*}[!htbp]
  \centering
  \small
  \setlength{\tabcolsep}{4pt}
  \caption{AUC@$3^\circ$ component ablations at $K=14$. Entries
  report averaged results. The final column
  averages the five dataset means. C: coarse grid; P: KV pooling;
  R: attention-output reuse.}
  \label{tab:components}
  \begin{tabular}{lrrrrrr}
    \toprule
    Configuration & DTU & ETH3D & 7-Scenes & ScanNet++ & nuScenes & Mean \\
    \midrule
    Official, $K=16$ & 83.21 & 65.97 & 13.93 & 79.48 & 43.41 & 57.20 \\
    \midrule
    C/P/R disabled & 84.08 & 63.77 & 13.85 & 78.69 & 43.18 & 56.71 \\
    C & 83.13 & 58.57 & 13.16 & 74.90 & 41.44 & 54.24 \\
    P & 83.34 & 62.23 & 13.74 & 77.86 & 42.48 & 55.93 \\
    R & 84.33 & 62.84 & 13.79 & 78.81 & 42.83 & 56.52 \\
    C+P & 82.96 & 56.63 & 12.88 & 73.56 & 41.85 & 53.58 \\
    C+R & 82.79 & 56.28 & 12.97 & 72.87 & 41.09 & 53.20 \\
    P+R & 84.18 & 58.85 & 13.55 & 77.23 & 42.49 & 55.26 \\
    C+P+R & 82.34 & 52.45 & 13.05 & 71.07 & 41.07 & 52.00 \\
    \bottomrule
  \end{tabular}
\end{table*}

\begin{table*}[!htbp]
  \centering
  \small
  \caption{Component accuracy--latency trade-offs. Each accuracy metric
  and latency is an unweighted mean over five datasets. Speedup
  divides the official 16-step reference mean latency by the
  configuration mean latency within this experiment.}
  \label{tab:component_tradeoffs}
  \begin{tabular}{lrrrrrr}
    \toprule
    Configuration & AUC@$3^\circ$ & AUC@$30^\circ$ & Inlier (\%) & AbsRel & Time (s) & Speedup \\
    \midrule
    Official, $K=16$ & 57.20 & 91.85 & 79.95 & 0.0310 & 1.804 & $1.00\times$ \\
    \midrule
    C/P/R disabled & 56.71 & 91.64 & 79.10 & 0.0317 & 1.629 & $1.11\times$ \\
    C & 54.24 & 91.15 & 77.74 & 0.0326 & 1.173 & $1.54\times$ \\
    P & 55.93 & 91.47 & 78.67 & 0.0322 & 1.410 & $1.28\times$ \\
    R & 56.52 & 91.65 & 79.33 & 0.0314 & 1.497 & $1.21\times$ \\
    C+P & 53.58 & 90.88 & 76.98 & 0.0335 & 0.932 & $1.94\times$ \\
    C+R & 53.20 & 90.95 & 77.02 & 0.0332 & 1.029 & $1.75\times$ \\
    P+R & 55.26 & 90.76 & 76.74 & 0.0352 & 1.266 & $1.42\times$ \\
    C+P+R & 52.00 & 90.61 & 76.04 & 0.0345 & 0.790 & $2.28\times$ \\
    \bottomrule
  \end{tabular}
\end{table*}

\subsection{Comparison across reconstruction models}
\label{app:external_comparison}

We report pose accuracy and latency for each dataset in \cref{tab:external_dataset}. For \cref{tab:external_macro} each dataset contributes equal weight, including for the geometry metrics.

The balanced tier obtains macro-average AUC@$3^\circ$ of 55.34 at
1.106 seconds, compared with 57.20 at 1.804 seconds for the official
16-step model. Their ratio of mean latencies is $1.63\times$.
The quality and fast tiers provide $1.31\times$ and $2.29\times$
speedup, with AUC values of 56.30 and 51.02, respectively. The
balanced tier has higher pose AUC and lower latency than the four
external models in the dataset macro-average.

Performance differs by dataset and output. On DTU, VGGT achieves
AUC@$3^\circ$ of 96.48, compared with 83.42 for our balanced tier.
On ScanNet++, $\pi^3$ achieves an inlier score of 94.34 and AbsRel
0.0143, compared with 91.72 and 0.0174 for the balanced tier, while
requiring 5.526 versus 1.894 seconds. The comparison therefore spans
both the backbone's reconstruction characteristics and the
accuracy--latency trade-off of the approximation schedule.

\begin{table*}[!htbp]
  \centering
  \small
  \setlength{\tabcolsep}{4pt}
  \caption{External comparison by dataset. Each pair reports
  AUC@$3^\circ$ (\%) and latency (seconds). Latencies are those
  of the evaluated implementations.}
  \label{tab:external_dataset}
  \begin{tabular}{lrrrrrrrrrr}
    \toprule
    Model & \multicolumn{2}{c}{DTU} & \multicolumn{2}{c}{ETH3D}
    & \multicolumn{2}{c}{7-Scenes} & \multicolumn{2}{c}{ScanNet++}
    & \multicolumn{2}{c}{nuScenes} \\
    \cmidrule(lr){2-3}\cmidrule(lr){4-5}\cmidrule(lr){6-7}\cmidrule(lr){8-9}\cmidrule(lr){10-11}
    & AUC & Time & AUC & Time & AUC & Time & AUC & Time & AUC & Time \\
    \midrule
    \dejaview ($K=16$) & 83.21 & 1.295 & 65.97 & 0.733 & 13.93 & 3.111 & 79.48 & 3.233 & 43.41 & 0.647 \\
    \dejaview ($K=14$) & 82.54 & 1.169 & 64.38 & 0.665 & 13.60 & 2.798 & 79.67 & 2.890 & 43.19 & 0.588 \\
    \ours, quality & 84.58 & 1.016 & 62.53 & 0.586 & 13.77 & 2.352 & 78.06 & 2.420 & 42.55 & 0.524 \\
    \ours, balanced & 83.42 & 0.843 & 60.36 & 0.494 & 13.39 & 1.853 & 77.43 & 1.894 & 42.10 & 0.448 \\
    \ours, fast & 80.27 & 0.631 & 49.98 & 0.383 & 12.83 & 1.273 & 71.19 & 1.293 & 40.84 & 0.354 \\
    VGGT & 96.48 & 3.007 & 35.42 & 1.707 & 11.13 & 7.246 & 14.92 & 7.570 & 39.15 & 1.525 \\
    FastVGGT & 87.03 & 2.899 & 17.55 & 1.785 & 9.73 & 5.826 & 12.10 & 5.741 & 30.58 & 1.763 \\
    AVGGT (VGGT) & 84.94 & 1.876 & 20.98 & 1.175 & 9.58 & 3.606 & 14.85 & 3.777 & 31.26 & 1.112 \\
    $\pi^3$ & 70.25 & 2.189 & 43.15 & 1.234 & 11.31 & 5.319 & 76.52 & 5.526 & 24.79 & 1.100 \\
    \bottomrule
  \end{tabular}
\end{table*}

\begin{table*}[!htbp]
  \centering
  \small
  \caption{Macro-average external comparison with equal weight for
  each dataset. AUC and inlier scores are percentages.
  These averages differ from scene-weighted
  summaries of the 153-scene evaluation set.}
  \label{tab:external_macro}
  \begin{tabular}{lrrrrr}
    \toprule
    Model & AUC@$3^\circ$ & AUC@$30^\circ$ & Inlier & AbsRel & Time (s) \\
    \midrule
    \dejaview ($K=16$) & 57.20 & 91.85 & 79.95 & 0.0310 & 1.804 \\
    \dejaview ($K=14$) & 56.67 & 91.72 & 79.56 & 0.0311 & 1.622 \\
    \ours, quality & 56.30 & 91.63 & 78.92 & 0.0317 & 1.380 \\
    \ours, balanced & 55.34 & 91.39 & 78.19 & 0.0323 & 1.106 \\
    \ours, fast & 51.02 & 90.54 & 75.89 & 0.0346 & 0.787 \\
    VGGT & 39.42 & 84.02 & 66.34 & 0.0457 & 4.211 \\
    FastVGGT & 31.40 & 81.85 & 58.90 & 0.0550 & 3.603 \\
    AVGGT (VGGT) & 32.32 & 80.50 & 57.86 & 0.0677 & 2.309 \\
    $\pi^3$ & 45.20 & 88.65 & 77.45 & 0.0334 & 3.073 \\
    \bottomrule
  \end{tabular}
\end{table*}

\subsection{Scene sensitivity, latency and memory}
\label{app:limits}

The results in \cref{sec:limits} use two different units of aggregation. The benchmark summary weights datasets equally. The win, tie and loss counts below give one vote to each scene.

\noindent\textbf{Scene-level outcomes.}
Against the official 16-step model, the quality, balanced, and
fast tiers obtain 49/21/83, 41/13/99, and 24/8/121 wins/ties/losses
in AUC@$3^\circ$ across 153 scenes, using an absolute difference below
0.2 points to define a tie. The balanced tier's scene-weighted mean
difference is $-1.61$ points. Its losses are particularly pronounced
on ETH3D, where it loses on 12 of 13 scenes.

\begin{figure*}[!htbp]
  \centering
  \includegraphics[width=0.68\textwidth]{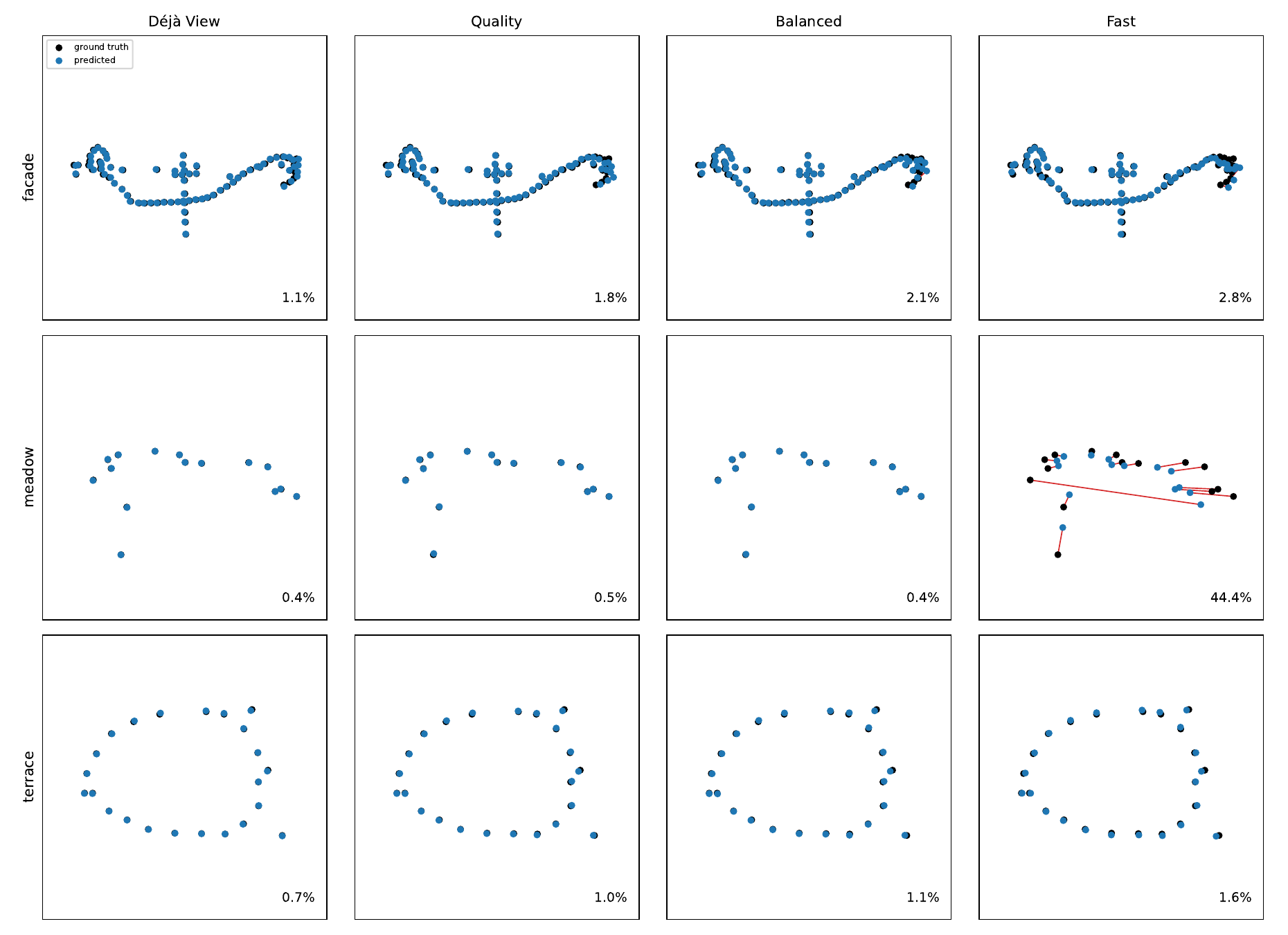}
  \caption{Camera centers on three ETH3D scenes, using all views of
  each scene, after similarity alignment to the ground truth and
  projected onto the dominant plane of the ground-truth cameras. Red
  segments connect each prediction to its ground truth. The number in
  each panel is the RMS position error as a percentage of the scene
  extent.}
  \label{fig:eth3d_traj}
\end{figure*}

\noindent\textbf{Camera trajectories on ETH3D.}
\Cref{fig:eth3d_traj} shows what these losses look like. We run each
model on all views of an ETH3D scene and align the predicted camera
centers to the ground truth with a similarity transform. Across the 13
scenes, the position error is 0.4--2.8\% of the scene extent for
\dejaview, 0.5--3.0\% for the quality tier, and 0.4--3.4\% for the
balanced tier. The losses of these two tiers are a small, uniform
increase of the camera error, and the trajectories remain intact. The
fast tier stays within 1.2--4.1\% on 12 scenes and breaks on
\texttt{meadow}, where several of the 15 cameras are misplaced.

\noindent\textbf{Scene characteristics.}
Grouping scenes by within-dataset all-view covisibility tertiles yields
balanced-tier AUC differences of $-2.50$, $-1.19$, and $-1.15$ points
for the low, middle, and high groups, respectively. Each group contains
51 scenes. For the fast tier, the corresponding differences are
$-7.25$, $-4.52$, and $-4.61$. Lower covisibility is associated with
larger losses in these measurements. For example, the fast tier
reduces AUC from 72.7 to 51.5 on ScanNet++ scene
\texttt{25f3b7a318}, which lies in the low-covisibility group.

\noindent\textbf{Scaling with the number of views.}
\Cref{tab:view_scaling} times each stage on ScanNet++ inputs of 24 to
300 views. The encoder and the prediction heads cost the same for all
configurations, and the loop dominates as the view count grows: it
takes 68\% of the \dejaview latency at 24 views and 93\% at 300 views.
The speedup of every tier therefore grows with the view count, from
$1.21\times$, $1.38\times$, and $1.69\times$ at 24 views to
$1.43\times$, $1.98\times$, and $3.40\times$ at 300 views for the
quality, balanced, and fast tiers.

For the benchmark settings in \cref{tab:compute_efficiency}, all three
tiers use 3.17 GiB versus 3.01 GiB for \dejaview in the standard
benchmark. Their peaks are 10.75 GiB versus 9.2 GiB on 300-view
ScanNet++ inputs and 12.62 GiB versus 10.8 GiB on 167--334-view
7-Scenes inputs.

\noindent\textbf{Where the additional memory goes.}
\Cref{tab:memory_modes} measures peak memory in a fresh process for
each configuration, with eight-frame decoder chunks. At 300 views,
the balanced tier with all step weights precomputed peaks at 12.61 GiB
against 8.49 GiB for \dejaview, an increase of 4.12 GiB. Merging the
weights when each step starts reduces the peak to 12.23 GiB. Also
releasing intermediate tensors as soon as they are last read reduces
it to 10.75 GiB. The model retains the multistep update history and the
cached attention output $A_t$ in all four execution modes.

Together, per-step merging and early release save 1.86 GiB at 300
views and preserve bit-identical outputs, leaving a 2.26 GiB increase
over \dejaview in this execution-mode comparison. Latency is nearly
unchanged: 11.85 seconds for the lean mode versus 11.86 seconds with
precomputed weights. At 24 views, both per-step and lean execution
take 0.356 seconds versus 0.349 seconds with precomputed weights.
Applying the low-rank adapters at run time without merging peaks at
14.47 GiB and takes 12.68 seconds at 300 views, about 7\% slower than
precomputed weights. At 24 views, its latency is about 21\% higher.

\begin{table*}[!htbp]
  \centering
  \small
  \setlength{\tabcolsep}{2pt}
  \caption{Latency in seconds against the number of views on ScanNet++
  inputs ($336\times504$). Encoder and heads are those of \dejaview;
  the tiers differ from them by at most 0.05 seconds. Speedup divides
  the total latency of \dejaview at $K=16$ by that of each tier.}
  \label{tab:view_scaling}
  \begin{tabular}{rrrrrrrrrrrrrr}
    \toprule
    & & & \multicolumn{4}{c}{Loop} & \multicolumn{4}{c}{Total}
      & \multicolumn{3}{c}{Speedup} \\
    \cmidrule(lr){4-7}\cmidrule(lr){8-11}\cmidrule(lr){12-14}
    Views & Encoder & Heads & \dejaview & Quality & Balanced & Fast
      & \dejaview & Quality & Balanced & Fast
      & Quality & Balanced & Fast \\
    \midrule
    24 & 0.034 & 0.100 & 0.293 & 0.223 & 0.179 & 0.123 & 0.428 & 0.354 & 0.310 & 0.254 & $1.21\times$ & $1.38\times$ & $1.69\times$ \\
    50 & 0.068 & 0.186 & 0.850 & 0.623 & 0.476 & 0.299 & 1.105 & 0.871 & 0.723 & 0.546 & $1.27\times$ & $1.53\times$ & $2.02\times$ \\
    100 & 0.133 & 0.354 & 2.707 & 1.917 & 1.399 & 0.804 & 3.197 & 2.393 & 1.874 & 1.281 & $1.34\times$ & $1.71\times$ & $2.50\times$ \\
    200 & 0.267 & 0.748 & 9.547 & 6.592 & 4.655 & 2.492 & 10.568 & 7.579 & 5.639 & 3.480 & $1.39\times$ & $1.87\times$ & $3.04\times$ \\
    300 & 0.401 & 1.102 & 20.661 & 13.996 & 9.744 & 5.038 & 22.109 & 15.451 & 11.182 & 6.494 & $1.43\times$ & $1.98\times$ & $3.40\times$ \\
    \bottomrule
  \end{tabular}
\end{table*}

\begin{table}[!htbp]
  \centering
  \small
  \setlength{\tabcolsep}{2pt}
  \caption{Peak GPU memory (GiB) and total latency (s) of the balanced
  tier under four ways of executing the adapters, on ScanNet++ inputs
  with eight-frame decoder chunks. Merged: all step weights
  precomputed, as timed in the paper. Run-time: low-rank path applied
  without merging. Per-step: weights merged when the step starts.
  Lean: per-step merging with early release of intermediate tensors.
  Per-step and lean give bit-identical outputs to merged.}
  \label{tab:memory_modes}
  \begin{tabular}{lrrrrrr}
    \toprule
    & \multicolumn{2}{c}{24 views} & \multicolumn{2}{c}{100 views}
      & \multicolumn{2}{c}{300 views} \\
    \cmidrule(lr){2-3}\cmidrule(lr){4-5}\cmidrule(lr){6-7}
    Execution & Peak & Time & Peak & Time & Peak & Time \\
    \midrule
    \dejaview ($K=16$) & 1.93 & 0.469 & 3.13 & 3.471 & 8.49 & 22.87 \\
    \midrule
    Merged & 2.45 & 0.349 & 4.78 & 2.121 & 12.61 & 11.86 \\
    Run-time & 2.07 & 0.422 & 5.15 & 2.401 & 14.47 & 12.68 \\
    Per-step & 2.07 & 0.356 & 4.40 & 2.118 & 12.23 & 11.85 \\
    Lean & 2.07 & 0.356 & 3.90 & 2.126 & 10.75 & 11.85 \\
    \bottomrule
  \end{tabular}
\end{table}

\section{Additional measurements and controls}
\label{app:figures_tables}

\subsection{Stepwise gains in the stacked and looped models}

\noindent\textbf{Attenuation within trajectories.}
Across the pooled steps of the matched-magnitude experiment, the median propagation gain is
0.968 for \dejaview, compared with 1.056, 1.021, and 1.215 for VGGT,
$\pi^3$, and DA3. The fractions of recorded steps with gain below one
are 66\%, 18\%, 36\%, and 0\%, respectively. These summaries quantify
attenuation within trajectories as well as the final retained fraction.
\Cref{app:precision} reports the dependence of these measurements on
numerical precision and perturbation amplitude.

\subsection{Finite perturbations and the gate-path control}
\label{app:extra_gates}

At 5\% of the native error magnitude, the reported cosine similarity
between first-order and finite-error propagation is 0.995. At native
magnitude, the cosine is 0.48 for step-12 injection and 0.20 for
step-7 injection. For step-12 injection, the predicted gain is 0.905
and the measured finite-error gain is 0.884. The local model captures
the aggregate gain closely in this setting, while the finite-error
direction changes substantially. The real direction has lower linear
gain than all four random directions in 64 of 65 scenes.

\noindent\textbf{Gate-path counterfactual.}
For each scene and propagated direction, we compute $G_t e_t$ and
$b_t=J_te_t-G_te_t$ from \cref{eq:gate_branch_split}. The first-step
gate response of the shuffled control matches that of the real error
to within $10^{-7}$. The full linear gains reproduce those of the Jacobian--vector products in \cref{app:gates} in all 65 scenes. Only 2 of 195 scene--step pairs
following step-12 injection have $\|J_te_t\|_2<\|G_te_t\|_2$;
none of the 520 pairs following step-7 injection do. Thus, in 713
of 715 pairs, the full response leaves a larger norm than the gate
path at the same incoming direction. These counts summarize repeated
steps within scenes.

\subsection{State and pose deviations at native perturbation magnitude}

\noindent\textbf{State attenuation and task deviation.}
We compare KV-pooling errors with eight random directions matched
to the error norm of each token, using the original 16-step model on
65 scenes with at most 32 views. At step 12, the structured perturbation leaves
a smaller final state deviation than all eight random controls in
62 scenes. Its rotation deviation exceeds the random-control median
in 50 scenes; both properties hold simultaneously in 47 scenes.
The median scene-wise state-deviation ratio is 0.75, while the median
rotation-deviation ratio is 1.26, with an interquartile range of
1.01--1.56. At step 7, structured perturbations leave smaller state
deviations in all 65 scenes, but exceed the random-control median
rotation deviation in only five. The median rotation ratio is 0.59,
with an interquartile range of 0.46--0.75.

The task measurements also vary with position. At step 12, structured
perturbations reduce AUC@$3^\circ$ against ground truth by 1.91 points,
compared with 1.51 points for the random controls, computed by taking
the median across directions within each scene and then averaging
across scenes. At step 7, the corresponding reductions are 0.59 and 2.52
points. A smaller final state deviation can therefore accompany
either a smaller or a larger task change, depending on the intervention.

\subsection{Task recovery in the adapted model}

The adapted 14-step model also exhibits position-dependent task costs.
KV pooling at step 2 ends with a median AUC@$3^\circ$ change of
$-0.1$ points, compared with $-0.8$ for step 7 and $-4.1$ for step 12.
Reuse at these positions ends at $-0.4$, $-1.4$, and $-3.5$ points,
respectively. These analysis scenes and view counts differ from the
official benchmark protocol used for the budget curves.

\subsection{Additional pooling in an adapted schedule}

\noindent\textbf{Sensitivity within an adapted schedule.}
The evaluated adapted 16-step configuration uses coarse-grid updates
at steps 1--8, KV pooling at steps 9--12, and exact, reuse, exact,
and reuse operations at steps 13--16. Reapplying reuse at steps 14
and 16 leaves the configured computation unchanged; these steps
therefore provide no additional reuse intervention. The remaining
exact global-attention evaluations are particularly sensitive:
replacing them with pooling at steps 13 and 15 reduces AUC@$3^\circ$
by 8.94 and 6.30 points, respectively; replacing them with reuse
reduces AUC by 3.52 and 4.62 points. This schedule exhibits a
concentration of sensitivity at its remaining exact attention steps.

\subsection{Dataset and metric dependence of the budget curves}

\Cref{tab:main} compares external models, the original backbone, and
our three operating points. The quality tier
uses 16 recurrent steps, while the balanced and fast tiers use 14.
Relative to the official 16-step model, the balanced tier provides
$1.44$--$1.71\times$ speedup across the five datasets. On DTU,
AUC@$3^\circ$ changes from 83.21 to 83.42, while on 7-Scenes it changes
from 13.93 to 13.39. The corresponding AbsRel values change from
0.0093 to 0.0095 and from 0.0349 to 0.0351. On ETH3D, the balanced
tier reduces AUC@$3^\circ$ by 5.61 points and increases AbsRel from
0.0268 to 0.0316, showing greater sensitivity to the approximations.

In the single-checkpoint sweep, increasing $K$ from 8 to 14 improves
macro-average AUC@$3^\circ$ from 42.35 to 52.00, while latency rises
from 0.719 to 0.790 seconds. At $K=16$, latency rises to 0.813 seconds
and AUC changes to 51.73, a decrease of 0.27 points. AUC@$30^\circ$ and the inlier score increase over this same
transition, while AbsRel increases by $0.1\times10^{-3}$. The budget
response therefore depends on the evaluation metric. Small reversals
also occur in the baselines: ScanNet++ AUC@$3^\circ$ decreases from
$K=14$ to $K=16$ by 0.19 points in the original model and by 0.14
in the plain baseline.

\subsection{Repeated failures on paired ScanNet++ inputs}

The paired view-count study in \cref{tab:long_sequence_placement}
identifies a subset of ScanNet++ scenes with large pooling-induced
AUC losses in the adapted model. These failures recur with 100-view
and 151--300-view inputs. Added early pooling overlaps the adapted
model's coarse-grid phase, making the existing operator schedule
part of the intervention context. Scene-level task metrics and the
operations already assigned to each step are therefore relevant when
transferring a placement rule to an adapted model.

\end{document}